\documentclass[letterpaper]{article} %
\usepackage{aaai26}
\usepackage{times}  %
\usepackage{helvet}  %
\usepackage{courier}  %
\usepackage[hyphens]{url}  %
\usepackage{graphicx} %
\usepackage{natbib}  %
\usepackage{caption} %
\usepackage{algorithm}
\usepackage{algorithmic}
\usepackage{multicol}
\usepackage{multirow}
\usepackage{booktabs}
\usepackage{pifont}
\usepackage{amsmath}
\usepackage{amsfonts}
\usepackage{listings}
\usepackage{tcolorbox}
\usepackage[utf8]{inputenc}
\usepackage{xcolor}
\newcommand{\cmark}{\ding{51}}
\newcommand{\xmark}{\ding{55}}
\newcommand{\modelname}[1]{3D-R1}
\usepackage{newfloat}
\usepackage{listings}
\DeclareCaptionStyle{ruled}{labelfont=normalfont,labelsep=colon,strut=off} %
\floatstyle{ruled}
\newfloat{listing}{tb}{lst}{}
\floatname{listing}{Listing}
\nocopyright 

\title{\raisebox{-0.5em}{\includegraphics[height=1.8em]{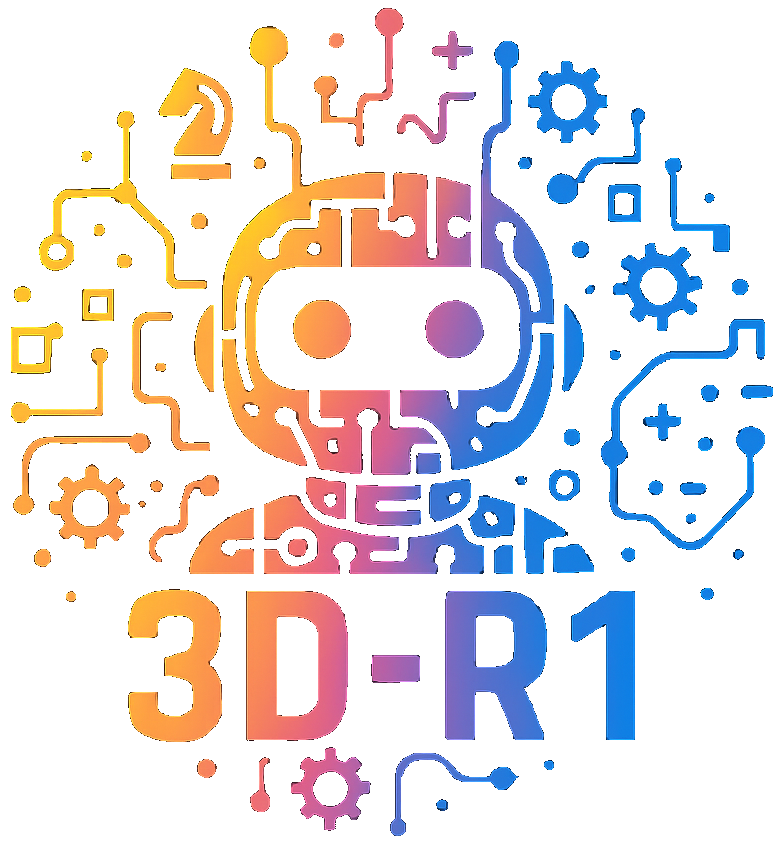}}~\modelname{}: Enhancing Reasoning in 3D VLMs for Unified Scene Understanding}

\author{
    Ting Huang$^{1*}$\quad Zeyu Zhang$^{2*\dag}$\quad Hao Tang$^{2\ddag}$
}

\affiliations{
    $^1$Shanghai University of Engineering Science\quad
    $^2$School of Computer Science, Peking University \vspace{0.1cm}\\
    \scriptsize $^*$Equal contribution. $^\dag$Project lead.
    $^\ddag$Corresponding author: bjdxtanghao@gmail.com.
}

\usepackage{bibentry}

\begin{document}

\makeatletter
\let\@oldmaketitle\@maketitle%
\renewcommand{\@maketitle}{\@oldmaketitle%
\vspace{-1cm}
\includegraphics[width=\linewidth]{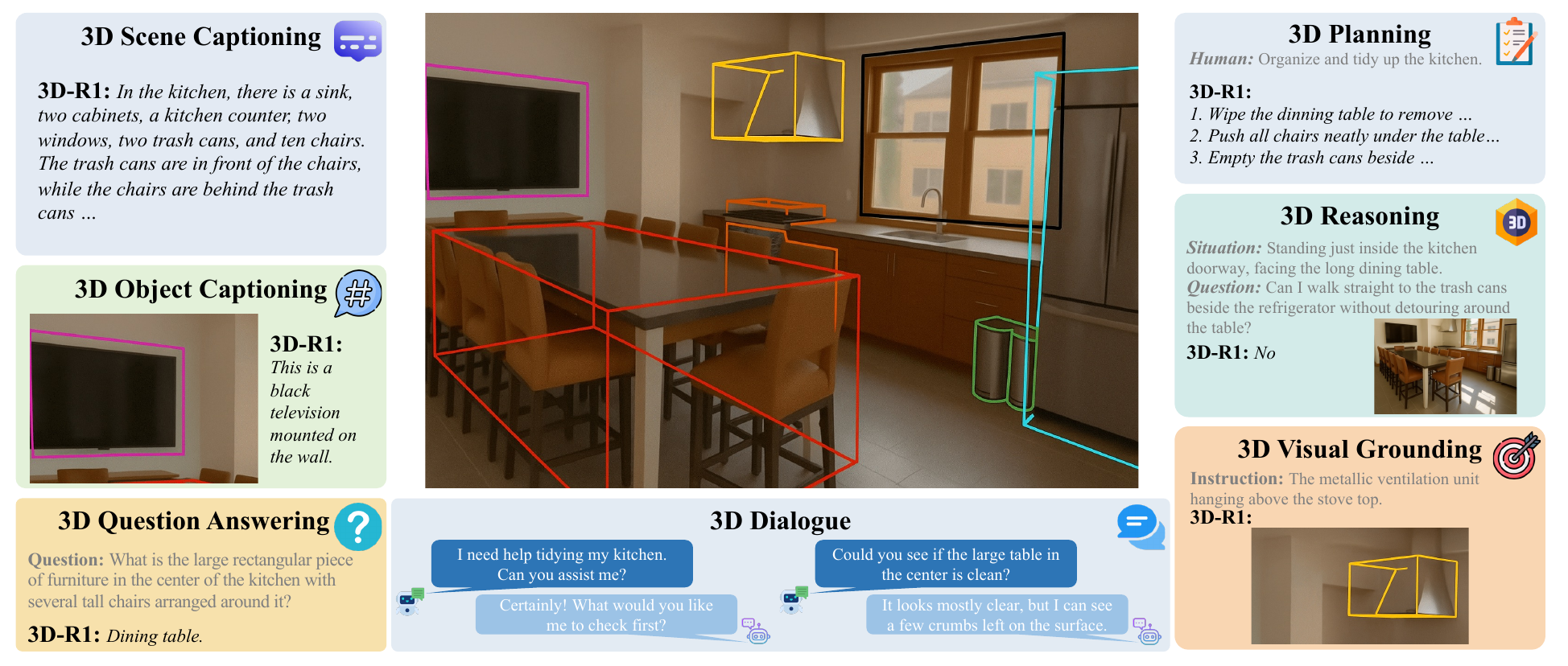}
\captionof{figure}{\textbf{\modelname{}} is an open-source generalist model that enhances the reasoning of 3D VLMs for unified scene understanding.
}\label{fig:main}\bigskip}%
\makeatother
\maketitle

\begin{abstract}
Large vision-language models (VLMs) have made significant strides in 2D visual understanding tasks, sparking interest in extending these capabilities to 3D scene understanding.
However, current 3D VLMs often struggle with robust reasoning and generalization due to limitations in high-quality spatial data and the static nature of viewpoint assumptions.
To address these challenges, we propose \textbf{\modelname{}}, a foundation model that enhances the reasoning capabilities of 3D VLMs.
Specifically, we first construct a high-quality synthetic dataset with CoT, named Scene-30K, leveraging existing 3D-VL datasets and a data engine based on Gemini 2.5 Pro. It serves as cold-start initialization data for \modelname{}.
Moreover, we leverage RLHF policy such as GRPO in the reinforcement learning training process to enhance reasoning capabilities and introduce three reward functions: a perception reward, a semantic similarity reward and a format reward to maintain detection accuracy and answer semantic precision.
Furthermore, we introduce a dynamic view selection strategy that adaptively chooses the most informative perspectives for 3D scene understanding.
Extensive experiments demonstrate that \modelname{} delivers an average improvement of 10\% across various 3D scene benchmarks, highlighting its effectiveness in enhancing reasoning and generalization in 3D scene understanding.
Code: \textcolor{blue}{\url{https://github.com/AIGeeksGroup/3D-R1}.} 
Website: \textcolor{blue}{\url{https://aigeeksgroup.github.io/3D-R1}}.
\end{abstract}

\begin{table*}[t!]
    \small
    \centering
    \caption{\textbf{Statistics of the public 3D-VL datasets that we draw on when synthesising the Scene‑30K dataset.} ``3D Scene / Obj.'' give the number of reconstructed scenes and annotated objects respectively. ``Task'' indicates the original benchmark focus,
    ``DC'' stands for Dense Captioning, ``QA'' for Question Answering, ``VG'' for Visual Grounding,
    and ``MT'' for Multi-tasking. 
    ``Anno.'' denotes language from human annotations and ``Syn.'' for template-based or LLM generated descriptions.}
    \resizebox{\linewidth}{!}{
        \begin{tabular}{l|cc|c|ccc|c|cc|r}
        \toprule
        \multirow{2}{*}{Dataset} & \multicolumn{2}{c|}{3D} & \multirow{2}{*}{Task} & Obj. & Scene & Obj. & Quality & \multirow{2}{*}{Anno.} & \multirow{2}{*}{Syn.} & \multirow{2}{*}{Total} \\
         & Scene & Obj.& & Caption & Caption & Referral & Check &  &  & \\
        \midrule
        ScanRefer~\cite{chen2020scanrefer}        & 800  & -    & DC\&VG & \xmark & \xmark & \cmark & \cmark & 52K  & - & 52K  \\
        Nr3D~\cite{achlioptas2020referit3d}       & 707  & -    & DC\&VG & \xmark & \xmark & \cmark & \cmark   & 42K  & 200K  & 242K \\
        ScanQA~\cite{azuma_2022_CVPR}             & 1.5K & 33K  & QA & - & - & - & \cmark  & 27K & - & 27K \\
        SceneVerse~\cite{jia2024sceneverse}       & 68K  & 1.5M & DC\&VG & \cmark & \cmark & \cmark & \cmark  & 190K & 2.3M & 2.5M   \\
        \midrule
        \textbf{Scene-30K}       & 1.5K  & 33K & MT & \cmark & \cmark & \cmark & \cmark  & - & 30K & 30K   \\
        \bottomrule
        \end{tabular}
    }
    \label{tab:stats_dataset}
\end{table*}

\section{Introduction}
3D scene understanding is a fundamental capability for intelligent systems, enabling a wide range of applications in embodied AI, robotics, and mixed reality~\cite{zhao2024openscan,song2025robospatial}. The ability of an agent to perceive and reason about 3D environments is crucial for tasks such as robotic manipulation, navigation, and long-horizon planning. Similarly, context-aware augmented and virtual reality applications require a rich semantic understanding of physical spaces to anchor virtual content and interactions in the real world. Furthermore, 3D scene understanding facilitates advanced spatial reasoning, such as interpreting spatial relations or inferring hidden object configurations, essential for agents to interact naturally with complex environments.

Researchers have recently extended vision-language models into the 3D domain to tackle tasks like 3D scene dense captioning (3D-DC)~\cite{scan2cap_2021,vote2cap2023,vote2cap++2024}, 3D object captioning~\cite{luo2024view}, 3D question answering (3D-QA)~\cite{azuma_2022_CVPR,bridgeqa2024}, 3D dialogue~\cite{LL3da2024,halacheva2025gaussianvlm}, 3D visual grounding (3D-VG)~\cite{jia2024sceneverse,huang2024chatscene}, and 3D reasoning and planning~\cite{halacheva2025gaussianvlm,LL3da2024}, as shown in Figure~\ref{fig:main}.
Current approaches typically employ either end-to-end modeling or leverage pretrained vision-language models (VLMs)~\cite{GPT4Scene,huang2024chatscene,xu2024pointllm}.

Despite this progress, current 3D vision language models still face significant limitations.
One of the primary challenges is enabling models to reason about complex spatial relationships and dynamic scene contexts. Traditional supervised fine-tuning (SFT) approaches often fail to effectively generalize across varied environments, as they are limited by the static nature of their training data and lack of adaptability.
Another limitation is the reliance on pre-defined views or representations. Several pipelines assume a fixed set of camera viewpoints or a global panoramic scene encoding, which can introduce irrelevant visual content and still miss critical details occluded in those views.

Recently, DeepSeek-R1~\cite{deepseekr12025} has successfully used reinforcement learning (RL) to induce large language models(LLMs) to autonomously emerge complex cognitive reasoning capabilities, begging our thinking to see whether we can leverage reinforcement learning (RL) to improve reasoning ability in 3D VLMs.

\begin{figure*}[t]
    \centering
    \includegraphics[width=\linewidth]{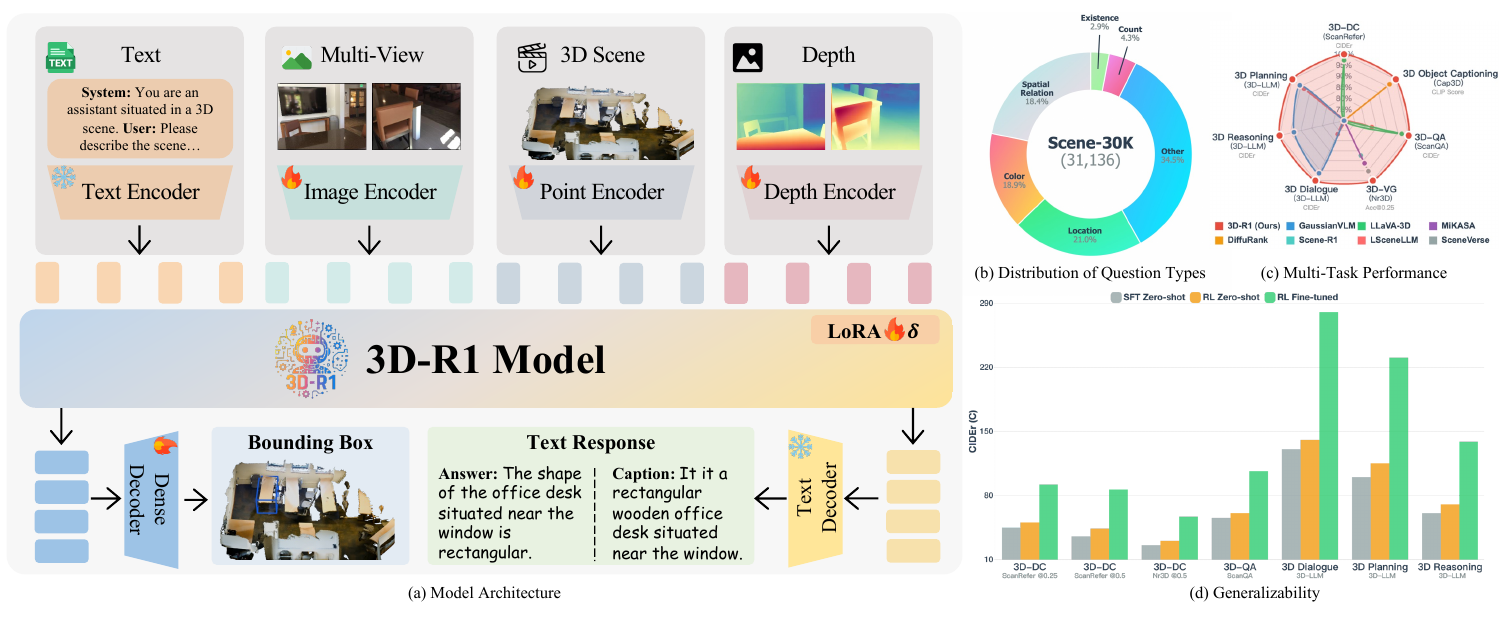}
    \caption{\textbf{(a) Architecture.} It takes text, multi-view images, 3D point clouds, and depth maps as input and formulates comprehensive 3D tasks as autoregressive sequence prediction.
    \textbf{(b) Distribution of question types.} Scene-30K contains diverse categories. 
    \textbf{(c) Multi-task performance.} \modelname{} demonstrates strong performance across various tasks. 
    \textbf{(d) Generalizability.} \modelname{} exhibits remarkable generalizability with enhanced reasoning capabilities.}
    \label{fig:pipeline}
\end{figure*}

To address these challenges, we propose \modelname{}, a foundation model to enhance reasoning capability in 3D scene understanding that integrates cold-start initialization with RL training.
First, we synthesize a high-quality 3D scene CoT dataset Scene-30K with diverse question types, as illustrated in Figure~\ref{fig:pipeline}(b).
Specifically, we design a 3D VLM to generate a concise textual description of a scene. This description captures objects, their relations, and their layout. The resulting textual descriptions are then passed to a reasoning model Gemini 2.5 Pro~\cite{geminipro2025} to produce high-quality CoT reasoning.
Finally, the dataset is refined through rule-based data filtering, ultimately obtaining a dataset with 30K complex CoT reasoning samples, which serves as the cold-start initialization dataset for \modelname{}.
Building on this foundation, we design a GRPO-based RLHF policy in the reinforcement learning fine-tune process and introduce three reward functions: a format reward, a perception reward, and a semantic similarity reward.
This process focuses on enhancing the model's reasoning capabilities while maintaining detection accuracy and answer semantic precision.
Furthermore, we introduce a dynamic view selection method, guiding the model learns to assign ranking scores to candidate viewpoints of the 3D scene and dynamically select the most informative views.
We conduct extensive experiments to enhance the capacities of reasoning within complex and diverse 3D environments. As shown in Figure~\ref{fig:pipeline}(c), \modelname{} achieves strong performance across various 3D scene tasks.

The main contributions of this work are as follows:
\begin{itemize}
    \item We introduce \textbf{\modelname{}}, a pioneering 3D VLM that leverages cold-start initialization and RL training to enhance reasoning capability in 3D scene understanding. Specifically, we design RLHF policy based on GRPO, including format, perception and semantic similarity reward function to improve reasoning in complex 3D scenes.
    \item A high-quality 30K scene CoT dataset is constructed to serve as a cold-start initialization data for 3D VLMs.
    Furthermore, we introduce dynamic view selection strategy that enables the model to dynamically select views of a 3D scene based on learned relevance scores.
    \item Extensive experiments demonstrate that \modelname{} achieves an average improvement of 10\% across various 3D scene benchmarks.
\end{itemize}

\section{Related Work}
\paragraph{3D scene understanding.}
3D scene understanding targets the comprehension of the semantic meaning of objects and their surrounding environment through the analysis of point clouds. 
In this study, we focus on several integral scene understanding tasks: 3D Scene Dense Captioning (3D-DC), 3D Object Captioning, 3D Question Answering (3D-QA), 3D Dialogue, 3D Visual Grounding (3D-VG), 3D Reasoning, and 3D Planning.
3D-DC involves producing descriptive language based on a 3D environment, encompassing both individual objects and the entire scene.
At the object level, models localize individual objects in a point cloud and describe each with natural language.
Scan2Cap~\cite{scan2cap_2021} first introduced this task by detecting objects in RGB-D scans and generating context-aware captions for each.
Subsequent work shifted from a two-stage ``detect-then-describe'' pipeline to an end-to-end transformer model.
For example, Vote2Cap-DETR~\cite{vote2cap2023} and its Vote2Cap-DETR++~\cite{vote2cap++2024} variant use a DETR-based encoder–decoder to jointly detect and caption objects in one pass.
At the scene level, models generate holistic descriptions of entire environments. The recent 3D-CoCa framework~\cite{huang20253d} integrated contrastive vision language pretraining with caption generation to produce semantically coherent scene descriptions~\cite{huang2025dc}. Likewise, LLM-augmented methods, such as LSceneLLM~\cite{zhi2024lscenellm} incorporated global context and language priors and used an LLM’s attention to focus on task-relevant areas and describe large cross-room scenes.

3D-QA extends the visual QA paradigm into 3D scenes, requiring spatial and cross-modal reasoning beyond 2D capabilities. The ScanQA~\cite{azuma_2022_CVPR} benchmark introduced this task by pairing 3D indoor scans with questions.
The follow-up work has increased the complexity, SQA3D~\cite{ma2022sqa3d}, for example, situated an embodied agent in the scene and poses questions about the agent’s surroundings, testing the model’s ability to interpret the agent’s viewpoint and reason about spatial relations in the 3D environment.

3D-VG focuses on locating referred objects in a 3D scene based on natural language expressions, requiring precise semantic and spatial alignment across modalities.
Recent research advances have explored unified transformer-based architectures and LLM-enhanced grounding. 3DVG-Trans~\cite{zhao2021_3DVG_Transformer} proposed a cross-modal transformer that fuses linguistic and point cloud level geometric features within a transformer-based framework.
Building on the capabilities of large language models, GPT4Scene~\cite{GPT4Scene} explored the zero-shot grounding setting. It integrated GPT-4 with 3D feature encoders via a lightweight alignment module and prompted the LLM to resolve spatial references from language alone.

Reinforcement learning (RL) techniques have recently been introduced to further improve multimodal 3D reasoning.
~\cite{chen2025compile} proposed to compile scene graphs with RL-enhanced MLLM, in a system called R1-SGG. They first train a multimodal LLM to output structured scene graphs from images and then refine it via RL with graph-centric rewards that promote high recall and semantic alignment of predicted objects and relationships.
In a related vein, ~\cite{dipr12025} introduced DIP-R1, an RL-based framework that guides a multimodal LLM to perform fine-grained visual inspection in complex scenes. These investigations showcase the potential of RL to improve 3D scene understanding in conjunction with large vision language models.
\begin{figure*}[t]
    \centering
    \includegraphics[width=\linewidth]{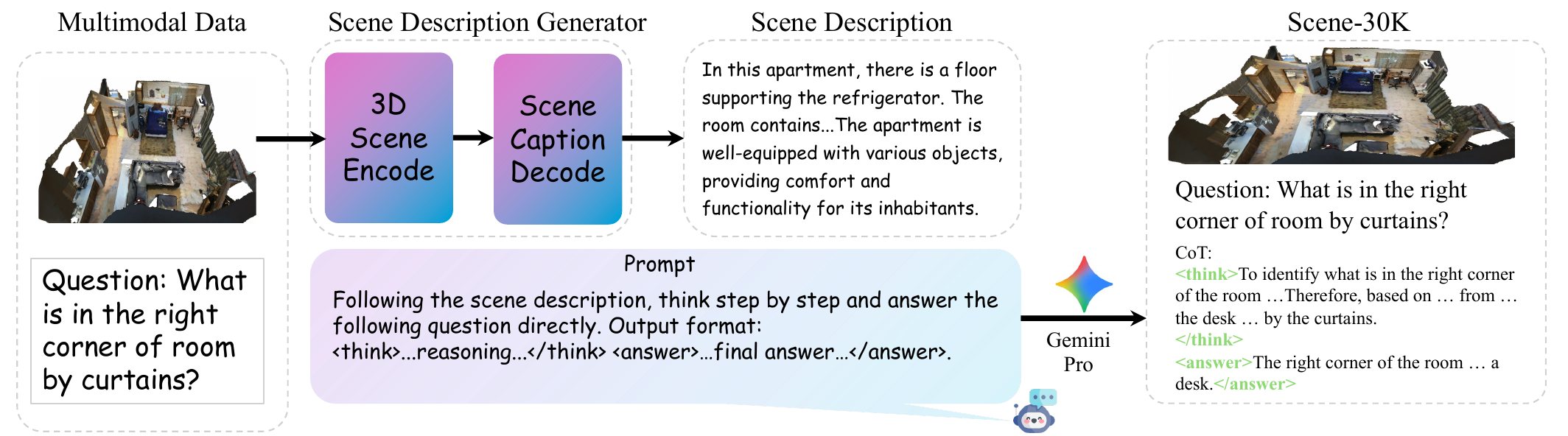}
    \caption{\textbf{CoT data engine.} The point cloud of a scene is first sent to scene dscription generator to get a description of the scene. Then based on the description, we apply Gemini 2.5 Pro to synthetic CoT data.}
    \label{fig:cold-start}
\end{figure*}

\paragraph{3D vision language models.}
Research on 3D vision–language models (3D-VLMs) has advanced rapidly, fueled by progress in large language models (LLMs).
The early 3D-VLMs focused on understanding 3D object point clouds~\cite{xu2024pointllm,minigpt3d2024}. PointLLM~\cite{xu2024pointllm} introduced an initial 3D-VLM that couples a point cloud encoder with an LLM, enabling the model to interpret colored object point clouds and answer questions about the shape and attributes of an object. Another line of work, MiniGPT-3D~\cite{minigpt3d2024} proposed an efficient strategy to align 3D data with language models utilizing 2D vision language priors.

More recently, researchers have shifted toward scene-level 3D-VLMs that can handle entire rooms or complex scenes with many objects.
For example, LLaVA-3D~\cite{zhu2024llava} augmented image patches with 3D position embeddings and performs joint 2D-3D instruction tuning, enabling the model to understand a whole scene and even output structured spatial information without relying on external detectors.
A recent work, 3D-LLaVA~\cite{deng20253dllava} takes a complementary approach, using a minimalist point-cloud-based pipeline with an integrated Omni Superpoint Transformer that acts as a visual encoder and multi-task decoder; this module selects salient 3D features, embeds interactive visual prompts, and can output grounded 3D segmentation masks, all within a single unified architecture.
\begin{figure*}[t]
    \centering
    \includegraphics[width=\linewidth]{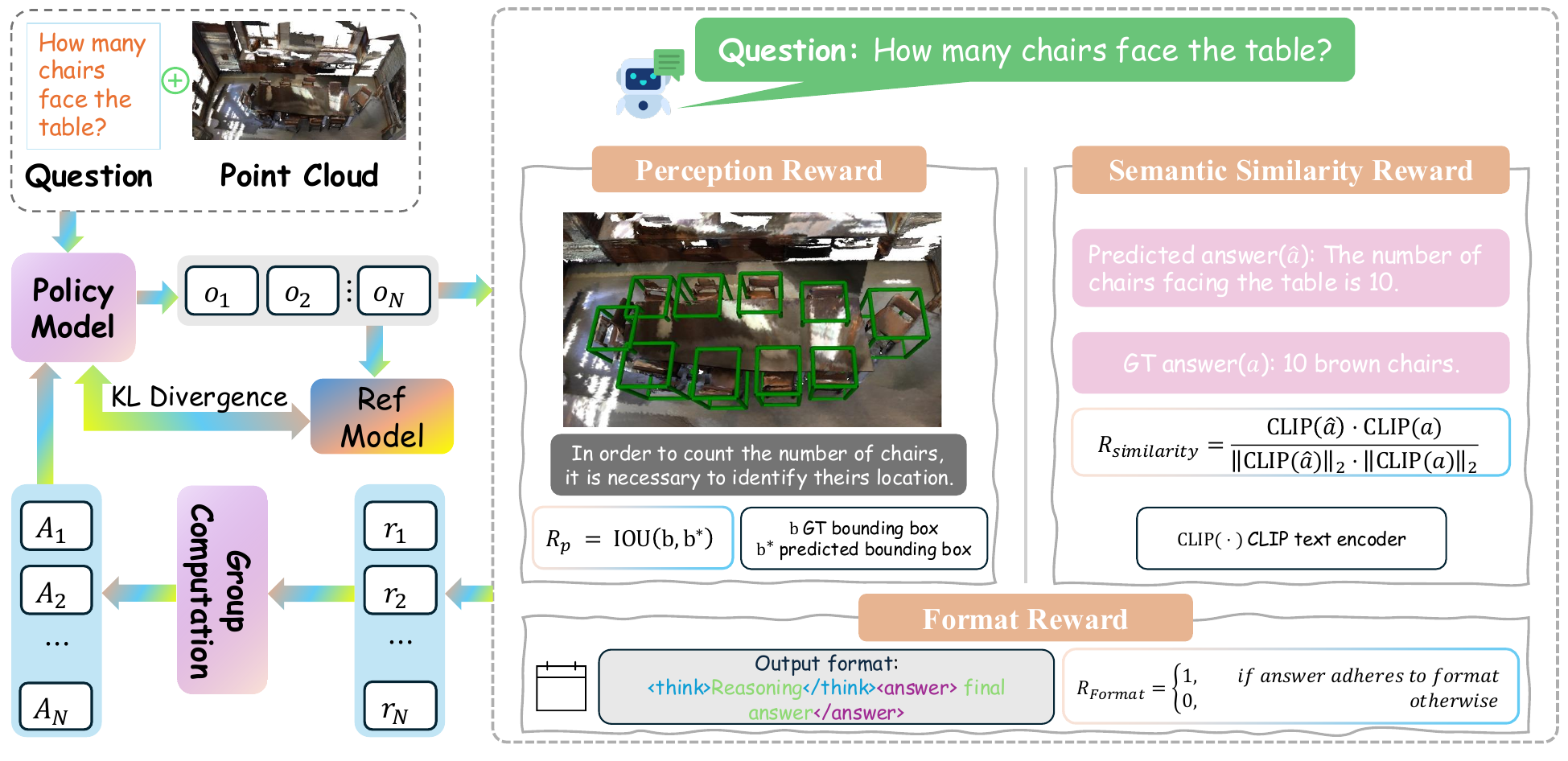}
    \caption{\textbf{The pipeline of Reinforcement Learning based GRPO.} The policy model generates $N$ outputs from a point cloud and question. Then perception IoU, semantic CLIP-similarity, and format-adherence rewards are computed, grouped, and combined with a KL term to a frozen reference model to update the policy.}
    \label{fig:grpo-pipeline}
\end{figure*}

\section{The Proposed Method}

\subsection{Overview}
The \modelname{} framework unfolds in two main phases.
In the first phase, we synthesize the Scene-30K dataset, which pairs 3D scenes with questions and coherent chains of thought (CoT).
In the second phase, we perform a cold start with the Scene-30K dataset to teach the base 3D VLM shown in Figure~\ref{fig:pipeline}(a) to reason in a ``human-like'' fashion.
Subsequently, as illustrated in Figure~\ref{fig:grpo-pipeline} we use RLHF policy such as Group Relative Policy Optimization (GRPO) and introduce two reward functions: a perception reward and a semantic similarity reward during the reinforcement learning training process to enhance the model's reasoning capabilities.
Finally, we introduce a dynamic view selection method that scores multiple candidate views of each 3D scene and adaptively chooses the most informative perspectives to answer the questions, ensuring the model focuses on relevant spatial context.

\subsection{CoT Data Engine}
We propose a CoT data engine for the construction of Chains of Thought (CoT)~\cite{wei2022chain} data tailored to 3D scene understanding.
This engine leverages the general reasoning capabilities of the large language model (LLM) to answer the questions with coherent, high-quality Chains of Thought (CoT).

As illustrated in Figure~\ref{fig:cold-start}, the point cloud of a 3D scene is fed into a scene description generator, which is a pre-trained 3D VLM that produces a concise textual summary of the scene. This summary captures objects, their relations, and their layout.
Then we design a comprehensive prompt that instructs Gemini 2.5 Pro~\cite{geminipro2025} to reason through the detailed logic structure to answer the question from the ScanQA~\cite{azuma_2022_CVPR} dataset. The prompt provides clear task instructions, specifies the required output format, and includes the previously generated scene description, guiding the model to produce structured step-by-step CoT reasoning.
Finally, the model outputs Chains of Thought (CoT) enclosed in \texttt{<think> … </think>} tags, followed by the final answer in \texttt{<answer> … </answer>} tags.
By running this pipeline on tens of thousands of 3D scenes and questions, we collect roughly 35K CoT examples, each containing a scene ID, a question, and the machine-generated \texttt{<think>} rationale and \texttt{<answer>} output.
Then these examples are refined through a rule-based filtering process that eliminates responses with missing structure or inconsistent reasoning; for more details, please see \textit{Appendix}. Finally, the 30K resulting examples constitute a high-quality CoT reasoning dataset, which we call Scene-30K dataset that serves as the cold-start initialization dataset for \modelname{}.

\subsection{Cold Start Stage}
Inspired by the success of DeepSeek-R1~\cite{deepseekr12025} in solving mathematical reasoning tasks through pure reinforcement learning, we first experiment with end-to-end RL training for our model, with the aim of inducing Chains of Thought (CoT) reasoning to answer the question solely from reward signals.
However, this approach proves highly unstable in the 3D VLM base model: the model frequently fails to generate coherent CoT sequences and, more critically, produces answers that are semantically misaligned.

To address the above issues, we adopt a cold start stage based on supervised fine-tuning on the Scene-30K dataset.
Leveraging the dataset, containing a question of scene, Chains of Thought (CoT) reasoning process, and corresponding final answer sequences, we fine-tune the 3D vision language model to bootstrap its ability to generate structured outputs in the form \texttt{<think>…</think><answer>…</answer>}.
This supervised initialization forces the model to learn the expected format for both the multistep reasoning process and the final answer, providing a stable and effective foundation for subsequent policy optimization with reinforcement learning (RL).

\subsection{Reinforcement Learning}
GRPO~\cite{deepseek-math} introduces an innovative approach rooted in reinforcement learning, showcasing impressive results in models such as DeepSeek R1~\cite{deepseekr12025}.
Its main objective is to improve the model’s reasoning skills by progressively improving its policy, using feedback from the precision of the responses sampled within a group.
\modelname{} decomposes the 3D scene understanding task into two distinct subtasks: scene perception and answer generation. It enables more focused learning and better generalization in complex 3D environments.
\paragraph{Policy samples.}
For a given input state $(x, q)$, where $x$ is the visual encoding of the input point cloud and $q$ is the textual encoding of the question, \modelname{} first generates $N$ distinct responses $\{o_1, o_1, \cdots, o_N\}$ from the current policy $\pi_\theta$.
To better guide policy learning and improve alignment between textual prompts and generated answers, we introduce a multi-reward mechanism.

\paragraph{Format reward.}
To ensure that the content generated by the model has a resolvable structure, we introduce Format Reward $R_{Format}$.
This reward detects through regularization expressions whether the generated results strictly follow the predefined format: \texttt{<think>Reasoning</think><Answer>final answer</Answer>}. The Format reward is defined as follows:
\begin{equation}
    R_{Format} =  \left\{\begin{matrix} 1 ,& \text{if Answer adheres to format}\\ 0 ,& \text{otherwise} \end{matrix}\right. .
\end{equation}

\paragraph{Perception reward.}
The perception reward focuses on the core objective of 3D scene perception: accurately identifying where the relevant objects' location is. It evaluates spatial precision by comparing the predicted bounding box $b^*$ with the ground-truth box $b$ using the intersection-over-union (IoU) metric.
By optimizing $R_p$, the model is encouraged to generate spatially precise and semantically grounded predictions that directly generate the correct answer. The Perception reward is defined as
\begin{equation}
    R_p = \text{IoU}(b,b^*).
\end{equation}

\paragraph{Semantic similarity reward.}
To encourage semantic coherence between the predicted answer $\hat{a}$ and the ground-truth answer $a$, we adopt a semantic similarity reward $R_{similarity}$. Specifically, we employ a pre-trained text encoder CLIP to obtain feature representations of both answers. The reward is computed as the cosine similarity between their embeddings:
\begin{equation}
    R_{similarity} = \frac{\mathrm{CLIP_{text}}(\hat{a})\cdot \mathrm{CLIP_{text}}(a)}{\left \| \mathrm{CLIP_{text}(\hat{a}) }  \right \| _2 \cdot \left \| \mathrm{CLIP_{text}(a)}  \right \|_2 }.
\end{equation}

\begin{table*}
    \caption{
        \textbf{3D scene dense captioning results on ScanRefer~\cite{chen2020scanrefer} and Nr3D~\cite{achlioptas2020referit3d}.}
        For fair comparison, we list methods that are trained under the standard per-word cross-entropy loss without additional 3D scenes.
        Our proposed \modelname{} surpasses previous 3D specialists on both datasets.
    }
    \centering
    \resizebox{\linewidth}{!}{
    \begin{tabular}{l|ccccccccc|cccc}
    \toprule
    \multirow{2}{*}{Method} & \multicolumn{9}{c|}{\large ScanRefer} & \multicolumn{4}{c}{\large Nr3D}      \\ 
    & C@0.25$\uparrow$ & B-4@0.25$\uparrow$ & M@0.25$\uparrow$ & R@0.25$\uparrow$ &  & C@0.5$\uparrow$ & B-4@0.5$\uparrow$ & M@0.5$\uparrow$ & R@0.5$\uparrow$ & C@0.5$\uparrow$ & B-4@0.5$\uparrow$ & M@0.5$\uparrow$ & R@0.5$\uparrow$ \\ \hline
    Scan2Cap~\cite{scan2cap_2021} & 56.82 & 34.18 & 26.29 & 55.27 & & 39.08 & 23.32 & 21.97 & 44.78 & 27.47 & 17.24 & 21.80 & 49.06 \\
    MORE~\cite{MORE_2022}         & 62.91 & 36.25 & 26.75 & 56.33 & & 40.94 & 22.93 & 21.66 & 44.42 & -     & -     & -     & -     \\
    SpaCap3D~\cite{spa2cap2022}   & -     & -     & -     & -     & & 44.02 & 25.26 & 22.33 & 45.36 & 33.71 & 19.92 & 22.61 & 50.50 \\
    REMAN~\cite{mao2023REMAN}     & 62.01 & 36.37 & 26.76 & 56.25 & & 45.00 & 26.31 & 22.67 & 46.96 & 34.81 & 20.37 & 23.01 & 50.99 \\
    D3Net~\cite{chen2021d3net}    & -     & -     & -     & -     & & 46.07 & 30.29 & 24.35 & 51.67 & 33.85 & 20.70 & 23.13 & 53.38 \\
    Contextual~\cite{zhong2022contextual3DdenseCap}& - & - & - & - & & 46.11 & 25.47 & 22.64 & 45.96 & 35.26 & 20.42 & 22.77 & 50.78\\
    UniT3D~\cite{unit3d2023}      & -     & -     & -     & -     & & 46.69 & 27.22 & 21.91 & 45.98 & -     & -     & -     & -     \\
    3DJCG~\cite{cai20223djcg}     & 64.70 & 40.17 & 27.66 & 59.23 & & 49.48 & 31.03 & 24.22 & 50.80 & 38.06 & 22.82 & 23.77 & 52.99 \\
    3D-VLP~\cite{jin2023-3D-VLP}  & 70.73 & 41.03 & 28.14 & 59.72 & & 54.94 & 32.31 & 24.83 & 51.51 & -     & -     & -     & -     \\
    3D-VisTA~\cite{zhu20233d-vista}& -    & -     & -     & -     & & 61.60 & 34.10 & 26.80 & 55.00 & -     & -     & -     & -     \\
    Vote2Cap-DETR~\cite{vote2cap2023} & 71.45 & 39.34 & 28.25 & 59.33 & & 61.81 & 34.46 & 26.22 & 54.40 & 43.84 & 26.68 & 25.41 & 54.43 \\
    LL3DA~\cite{LL3da2024}            & 74.17 & 41.41 & 27.76 & 59.53 & & 65.19 & 36.79 & 25.97 & 55.06 & 51.18 & 28.75 & 25.91 & 56.61 \\
    Vote2Cap-DETR++~\cite{vote2cap++2024}& 76.36 & 41.37 & 28.70 & 60.00 & & 67.58 & 37.05 & 26.89 & 55.64 & 47.08 & 27.70 & 25.44 & 55.22 \\
    LEO~\cite{huang2024an}         & -    & -     & -     & -     & & 72.40 & 38.20 & 27.90 & 58.10 & -     & -     & -     & -     \\
    ChatScene~\cite{huang2024chatscene}  & -     & -     & -     & -     & & 77.20 & 36.30 & 28.00 & 58.10 & -     & -     & -     & -     \\
    LLaVA-3D~\cite{zhu2024llava}         & -     & -     & -     & -     & & 84.10 & 42.60 & 29.00 & 63.40 & -     & -     & -     & -     \\
    BiCA~\cite{bica2025}          & 78.42 & 41.46 & 28.82 & 60.02 & & 68.46 & 38.23 & 27.56 & 58.56 & 48.77 & 28.35 & 25.60 & 55.81 \\
    3D CoCa~\cite{huang20253d}    & 85.42 & 45.56 & 30.95 & 61.98 & & 77.13 & 41.23 & 28.52 & 57.40 & 52.84 & 29.29 & 25.55 & 56.43 \\
    3D-LLaVA~\cite{deng20253dllava}& -    & -     & -     & -     & & 78.80 & 36.90 & 27.10 & 57.70 & -     & -     & -     & -     \\
    Spatial 3D-LLM~\cite{spatial3dllm}& - & -     & -     & -     & & 72.20 & 34.60 & 23.10 & 54.30 & -     & -     & -     & -     \\
    \midrule
    \textbf{\modelname{} (Ours)} & \textbf{91.85} & \textbf{48.76} & \textbf{32.14} & \textbf{62.23} & & \textbf{86.45} & \textbf{44.34} & \textbf{29.78} & \textbf{64.50} & \textbf{56.98} & \textbf{31.13} & \textbf{26.12} & \textbf{57.54}    \\
    \bottomrule
    \end{tabular}
    }
    \label{exp:comparison_on_densecaption}
\end{table*}

\paragraph{Policy update.}
Inspired by Group Relative Policy Optimization (GRPO)~\cite{deepseek-math}, we select multiple responses from the current policy as candidate responses.
Each output is assigned a scalar reward, resulting in a reward vector $\textbf{\text{r}} = \{ r_1, r_2, \cdots , r_N \}$, computed by task-specific reward functions that evaluate the quality of each output.
To assess the quality of each response relative to others, we normalize the rewards by computing the mean and standard deviation:
\begin{equation}
    \hat{A}_i = \frac{r_i - \mathrm{mean}(\textbf{\text{r}})}{\text{std}(\textbf{\text{r}})},
\end{equation}
where $\hat{A}_i$ denotes the advantage of the $i$-th response. These advantages are then used to update the policy by maximizing the following clipped objective:
\begin{equation}
\allowdisplaybreaks
\begin{aligned}
\mathcal{J}_{\mathrm{GRPO}}(\theta) 
= &\mathbb{E}_{c}
\Biggl[ \frac{1}{G}\sum_{i=1}^G \biggl( 
\min\left(\frac{\pi_\theta(o_i|q)}{\pi_{\theta_{\mathrm{old}}}(o_i|q)}\hat{A}_i,\right.\\[5pt]
&\quad\left.\mathrm{clip}\left(
\frac{\pi_\theta(o_i|q)}{\pi_{\theta_{\mathrm{old}}}(o_i|q)},
1-\varepsilon,
1+\varepsilon
\right)\hat{A}_i\right)\\[5pt]
&\quad -\beta \cdot \mathbb{D}_{\mathrm{KL}}(\pi_\theta\|\pi_{\mathrm{ref}})\biggr)\Biggr].
\end{aligned}
\end{equation}

\subsection{Dynamic View Selection}
To bridge the gap between 3D scene representations and the 2D perspective inputs that VLMs expect, we introduce a dynamical view selection module. The core idea is to automatically select a set of informative 2D views from a 3D scene that best convey the content of the scene to the vision-language model.

\paragraph{Candidate view generation.}
For each 3D scene, we first generate a pool of candidate views. We use the 3D point cloud to render RGB images from various viewpoints. In practice, we sample camera positions uniformly around the scene or at strategic locations to obtain a diverse set of perspective images. Each candidate view is processed by a pre-trained visual encoder to extract features. This pre-trained model provides a rich description of the view content without any additional 3D training, capitalizing on the learned 2D visual semantics.

\paragraph{View scoring metrics.}
We design three complementary scoring functions to evaluate each candidate view with respect to a given textual context. These scores are used to prioritize critical and diverse views. Specifically, for each scene $v$ and input text $t$, we calculate $S_{\mathrm{Text} \to \mathrm{3D} }$, $S_{\mathrm{Image} \to \mathrm{3D} }$, and $S_{\mathrm{CLIP}}$ as follows:
\begin{equation}
\begin{aligned}
    S_{\mathrm{Text}\to\mathrm{3D}}(v,t) &= \frac{E_{\text{text}}(t) \cdot E_{\text{3D}}(v)}{\left\lVert E_{\text{text}}(t)\right\rVert_2 \left\lVert E_{\text{3D}}(v)\right\rVert_2} \\
    S_{\mathrm{Image}\to\mathrm{3D}}(v,t) &= \frac{1}{|I(t)|}\sum_{i\in I(t)} \frac{E_{\text{img}}(i) \cdot E_{\text{3D}}(v)}{\left\lVert E_{\text{img}}(i)\right\rVert \left\lVert E_{\text{3D}}(v)\right\rVert} \\
    S_{\mathrm{CLIP}}(v,t) &= \frac{E_{\text{CLIP}}^{\text{txt}}(t) \cdot E_{\text{CLIP}}^{\text{img}}(R(v))}{\left\lVert E_{\text{CLIP}}^{\text{txt}}(t) \right\rVert \left\lVert E_{\text{CLIP}}^{\text{img}}(R(v)) \right\rVert},
\end{aligned}
\end{equation}
where
$E_{\text{text}}(\cdot)$ denotes text encoder,
$E_{\text{img}}(\cdot)$ denotes image encoder,
$E_{\text{3D}}(\cdot)$ denotes point encoder,
$I(t)$ is the set of multi-view images of the scene,
$R(v)$ renders scene $v$ into 2D image,
$E_{\text{CLIP}}^{\text{txt}}(\cdot)$ and $E_{\text{CLIP}}^{\text{img}}(\cdot)$ are the text and image branches of CLIP,
and $\lVert\cdot\rVert$ is the Euclidean norm.

\paragraph{Dynamic score fusion.}
The above scores are combined to produce an overall utility score for each view $U(v)$. Instead of manually tuning their relative importance, we dynamically learn the weight of these components. We introduce learnable parameters $w_t$, $w_c$, $w_{clip}$ for the text relevance, coverage, and CLIP alignment scores, respectively. This adaptive fusion ensures that $U(v)$ emphasizes the most useful views for each scenario. $U(v)$ is defined as follows:
\begin{equation}
    U(v) = w_t \cdot S_{\mathrm{Text} \to \mathrm{3D}} + w_c \cdot S_{\mathrm{Image} \to \mathrm{3D}} + w_{clip} \cdot S_{\mathrm{CLIP}},
\end{equation}
where $w_c+ w_{clip}=1$, $w_t$ as an independent scalar. This allows the model to dynamically adjust the influence of textual grounding relative to visual signals. To stabilize training, we apply an L2 regularization term on $w_t$, encouraging it to stay near a target value (e.g., $\mu$ = 0.3), which prevents overly dominant text influence.

\begin{table*}[htbp]
    \caption{
        \textbf{3D question answering results on ScanQA~\cite{azuma_2022_CVPR}.}
        \modelname{} out-performs previous methods on the validation set and two test sets.
    }
    \centering
    \resizebox{\linewidth}{!}{
    \begin{tabular}{lcccccccccccccc}
    \toprule
    \multirow{2}{*}{Method} & \multicolumn{4}{c}{Validation} & & \multicolumn{4}{c}{Test w/ object} & & \multicolumn{4}{c}{Test w/o object}\\ 
    \cline{2-5} \cline{7-10} \cline{12-15}
    & C$\uparrow$ & B-4$\uparrow$ & M$\uparrow$ & R$\uparrow$ & & C$\uparrow$ & B-4$\uparrow$ & M$\uparrow$ & R$\uparrow$ & & C$\uparrow$ & B-4$\uparrow$ & M$\uparrow$ & R$\uparrow$ \\ \hline
    ScanQA~\cite{azuma_2022_CVPR} & 64.86 & 10.08 & 13.14 & 33.33 & & 67.29 & 12.04 & 13.55 & 34.34 &  & 60.24 & 10.75 & 12.59 & 31.09 \\
    Clip-Guided~\cite{parelli2023clip-guided} & - & - & - & - & & 69.53 & 14.64 & 13.94 & 35.15 & & 62.83 & 11.73 & 13.28 & 32.41 \\
    3D-VLP~\cite{jin2023-3D-VLP}              & 66.97 & 11.15 & 13.53 & 34.51 & & 70.18 & 11.23 & 14.16 & 35.97 & & 63.40 & 15.84 & 13.13 & 31.79 \\
    3D-LLM~\cite{hong2023dllm}                & 69.40 & 12.00 & 14.50 & 35.70 & & 69.60 & 11.60 & 14.90 & 35.30 & & -     & -     & -     & -     \\
    3D-VisTA~\cite{zhu20233d-vista}           & 69.60 & 10.40 & 13.90 & 35.70 & & 68.60 & 10.50 & 13.80 & 35.50 & & 55.70 & 8.70  & 11.69 & 29.60 \\
    LL3DA~\cite{LL3da2024}                    & 76.79 & 13.53 & 15.88 & 37.31 & & 78.16 & 13.97 & 16.38 & 38.15 & & 70.29 & 12.19 & 14.85 & 35.17 \\
    BridgeQA~\cite{bridgeqa2024}              & -     & -     & -     & -     & & 83.75 & 24.06 & 16.51 & 43.26 & & 79.34 & 17.74 & 15.60 & 41.18 \\
    ChatScene~\cite{huang2024chatscene}       & 87.70 & 14.30 & 18.00 & 41.60 & & -     & -     & -     & -     & & -     & -     & -     & -     \\
    3D-LLaVA~\cite{deng20253dllava}           & 92.60 & 17.10 & 18.40 & 43.10 & & -     & -     & -     & -     & & -     & -     & -     & -     \\
    Scene-LLM~\cite{scenellm2025}             & 80.00 & 12.00 & 16.60 & 40.00 & & -     & -     & -     & -     & & -     & -     & -     & -     \\
    Spatial 3D-LLM~\cite{spatial3dllm}        & 82.50 & 13.90 & 16.80 & 39.10 & & -     & -     & -     & -     & & -     & -     & -     & -     \\
    LSceneLLM~\cite{zhi2024lscenellm}         & 88.24 & -     & 17.95 & 40.82 & & -     & -     & -     & -     & & -     & -     & -     & -     \\
    LEO~\cite{huang2024an}                    & 101.40& 13.20 & 20.00 & 49.20 & & -     & -     & -     & -     & & -     & -     & -     & -     \\
    LLaVA-3D~\cite{zhu2024llava}              & 103.10& 16.40 & 20.80 & 49.60 & & -     & -     & -     & -     & & -     & -     & -     & -     \\
    GaussianVLM~\cite{halacheva2025gaussianvlm}& -    & -     & \textbf{22.90} & 34.80 & & -    & -     & -     & - & & -     & -     & -    & -     \\
    \midrule
    \textbf{\modelname{} (Ours)}               & \textbf{106.45} & \textbf{17.80} & 22.13 & \textbf{51.23} & & \textbf{94.65} & \textbf{35.34} & \textbf{\textbf{27.34}} & \textbf{54.35} & & \textbf{89.56} & \textbf{26.34} & \textbf{27.34} & \textbf{52.38}    \\
    \bottomrule
    \end{tabular}
    }
    \label{exp:comparison_on_questionanswer}
\end{table*}

\section{Experiment}

\subsection{Datasets and Metrics}
\paragraph{Datasets.}
To obtain the cold-start dataset, as shown in Tab~\ref{tab:stats_dataset}, we use ScanQA~\cite{azuma_2022_CVPR}, ScanRefer~\cite{chen2020scanrefer}, Nr3D~\cite{achlioptas2020referit3d} and SceneVerse~\cite{jia2024sceneverse} datasets to synthesize the Scene-30K dataset.
In downstream tasks, we incorporate standard benchmarks including ScanRefer~\cite{chen2020scanrefer} and Nr3D~\cite{achlioptas2020referit3d} dataset for 3D-DC and 3D-VG, Cap3D~\cite{luo2023scalable} for 3D object captioning, ScanQA~\cite{azuma_2022_CVPR} dataset for 3D-QA , 3D-LLM~\cite{hong2023dllm} for 3D dialogue and planning and SQA3D~\cite{ma2022sqa3d} for 3D reasoning.

\paragraph{Metrics.}
For 3D-VG task, we use metric Acc@$s$IoU, which reports grounding accuracy with different IoU scores $s$ between the predicted and ground truth bounding boxes.
For the 3D object captioning task, we adopt both human and automated evaluation metrics. Human evaluation involves A/B testing to assess two key aspects: caption quality and hallucination rate, reporting average preference scores and win/loss rates. For automated evaluation, we follow CLIP-based retrieval metrics, including cosine similarity scores and retrieval precision~\cite{cliprpercision} at top-1, top-5 and top-10 (R@1, R@5, R@10).
For 3D-DC, 3D-QA, 3D dialogue, 3D reasoning and 3D planning tasks, we use the metrics CIDEr~\cite{cider2015}, BLEU~\cite{bleu2002}, METEOR~\cite{meteor2005} and ROUGE-L~\cite{rouge2004}, which are briefly denoted by C, B-4, M and R, respectively, to evaluate the quality of the generated textual responses.

\begin{table*}[t]
\caption{\textbf{3D object captioning results} on Cap3D~\cite{luo2023scalable}. All A/B testing represents captions from other methods vs. ours.
$\dag$ indicates DiffuRank~\cite{luo2024view} trained with top 6 views.}
\resizebox{\linewidth}{!}{
\begin{tabular}{lcccccccccccc}
\toprule
\multirow{2}{*}{Method} & \multicolumn{3}{c}{Quality A/B test} & & \multicolumn{3}{c}{Hallucination A/B test} & & \multicolumn{4}{c}{CLIP} \\
\cline{2-4} \cline{6-8} \cline{10-13}
 & Score(1-5) & Win \% & Lose \%  & & Score(1-5) & Win \% & Lose \% & & Score & R@1 & R@5 & R@10\\
\midrule
Cap3D~\cite{luo2023scalable} & 2.62 & 32.70 & 60.20 & & 2.43 & 25.80 & 63.90 & & 71.20 & 20.50 & 40.80 & 51.90 \\
DiffuRank (Allviews 28-views) & 2.91 & 37.90 & 43.60 & & 2.85 & 35.10 & 47.20 & & 73.50 & 24.90 & 46.70 & 55.70 \\
DiffuRank (Horizontal 6-views) & 2.84 & 35.20 & 44.50 & & 2.90 & 36.20 & 40.90 & & 73.80 & 25.80 & 46.70 & 55.90 \\
DiffuRank (Bottom 6-views) & 2.74 & 31.10 & 52.00 & & 2.61 & 30.10 & 57.00 & & 72.80 & 4.60 & 45.10 & 55.20 \\
DiffuRank~\cite{luo2024view}$\dag$ & - & - & - & & - & - & - & & 74.60 & 26.70 & 48.20 &57.50   \\
\midrule
\textbf{\modelname{} (Ours)} & \textbf{4.32} & \textbf{34.56} & \textbf{65.34} & & \textbf{4.21} & \textbf{27.34} & \textbf{69.12} & &\textbf{77.34}& \textbf{32.23} & \textbf{55.45} &\textbf{63.12}   \\
\bottomrule
\end{tabular}
}
\label{exp:comparison_on_objectcaption}
\end{table*}

\begin{table*}[ht]
\centering
\caption{\textbf{3D dialogue and planning} results on 3D-LLM~\cite{hong2023dllm}. \textbf{3D reasoning} results on SQA3D~\cite{ma2022sqa3d}.}
\resizebox{\textwidth}{!}{
\begin{tabular}{lccccccccccccccccccc}
\toprule
\multirow{2}[2]{*}{Method} & \multicolumn{4}{c}{\textbf{Dialogue}} & &\multicolumn{4}{c}{\textbf{Reasoning}} & & \multicolumn{4}{c}{\textbf{Planning}} \\
\cline{2-5} \cline{7-10} \cline{12-15}
 & C$\uparrow$ & B-4$\uparrow$ & M$\uparrow$ & R$\uparrow$& & C$\uparrow$ & B-4$\uparrow$ & M$\uparrow$ & R$\uparrow$ & & C$\uparrow$ & B-4$\uparrow$ & M$\uparrow$ & R$\uparrow$ \\
\toprule
LL3DA~\cite{LL3da2024} & 190.01 & 23.95 & 23.50 & 40.61 & & - & - & - & - &  & 128.80 & 12.95 & 17.05 & 39.25 \\
Spatial 3D-LLM~\cite{spatial3dllm} & -  & - & -  & - & & -  & - & -  & - & & 195.92 & 14.65 & 18.95 & 36.93 \\
LSceneLLM~\cite{zhi2024lscenellm} & 104.98 & - & 21.26 & 36.00 & & - & - & - & - & & 214.63 & - & 21.05 & 47.05 \\
LEO~\cite{huang2024an}            & -      & - & -     & -     & & 124.70 & 9.40 & 25.50 & 48.40 & & - & - & - & - \\
GPT-4o~\cite{openai2024gpt4technicalreport} & 200.34 & 26.47 & 26.35 & 47.88 & & 120.45 &19.34 &25.45 & 49.34& & 210.23 &18.67 &42.23& 45.23 \\
Gemini 2.5 Pro~\cite{geminipro2025} & 210.23 & 27.34 & 28.12 & 48.22 & & 125.23 &20.23 &27.34 & 55.34& & 215.34 &20.19 &44.34& 46.23 \\
GaussianVLM~\cite{halacheva2025gaussianvlm} &270.10 & 31.50 & 55.70 & 48.60 & & 129.60 & 17.10 & 26.40 & 50.20 & & 220.40 & 20.30 & 44.50 & 48.00 \\
\midrule
\textbf{\modelname{} (Ours)} & \textbf{280.34} & \textbf{39.45} & \textbf{66.89} & \textbf{55.34} & & \textbf{138.67} & \textbf{23.56} & \textbf{35.45} & \textbf{60.02} & &\textbf{230.50} & \textbf{25.45} & \textbf{48.34} & \textbf{55.67} \\
\bottomrule
\end{tabular}
}
\label{exp:comparison_on_dialoguereasoningplanning}
\end{table*}

\subsection{Main Results}
We evaluate the model's capacity to understand and reason in 3D environments via 3D-DC, 3D object captioning, 3D-QA, 3D dialogue, 3D-VG, 3D reasoning, and 3D planning.

\paragraph{3D scene dense captioning.}
It demands a model to localize and describe an object in a 3D scene. We compare SOTA methods on the widely used ScanRefer~\cite{chen2020scanrefer} and Nr3D~\cite{achlioptas2020referit3d} benchmarks. The results in Table~\ref{exp:comparison_on_densecaption} show that our method consistently outperforms existing methods on both datasets.

\paragraph{3D object captioning.}
This task requires the model to describe a localized object in a 3D scene. We compare SOTA methods on Cap3D~\cite{luo2023scalable} benchmark. As shown in Table~\ref{exp:comparison_on_objectcaption}, ``Allviews 28-views'' indicates DiffuRank~\cite{luo2024view} trained with all 28 views, ``Horizontal 6-views'' with 6 horizontal views, ``Bottom 6-views'' with 6 bottom views. The results show that \modelname{} achieves the highest scores across all evaluation criteria.

\paragraph{3D question answering.}
It requires a model to generate responses to the natural language queries questioning towards a 3D scene. We compare SOTA methods on the ScanQA~\cite{azuma_2022_CVPR} validation set as well as two test benchmarks in Table~\ref{exp:comparison_on_questionanswer}. The results show that our method consistently outperforms existing methods on all evaluation sets.

\paragraph{3D visual grounding.}
It requires a model to accurately localize the object referenced by a natural language expression within a 3D scene. We benchmark state-of-the-art methods on the widely used Nr3D~\cite{achlioptas2020referit3d} and ScanRefer~\cite{chen2020scanrefer} datasets as seen in Table~\ref{exp:comparison_on_vg}. We can see that our method consistently outperforms existing methods on both datasets.

\paragraph{3D reasoning.}
It requires the model to infer spatial or functional relationships between objects based on contextual cues within a 3D scene. We evaluate on the SQA3D~\cite{ma2022sqa3d} benchmark and report standard metrics in Table~\ref{exp:comparison_on_dialoguereasoningplanning}. The results show that \modelname{} achieves the highest scores across all metrics.

\paragraph{3D dialogue.}
This task involves generating interactive context-aware responses grounded in the 3D scene. We compare our method on the 3D-LLM~\cite{hong2023dllm} dataset, as shown in Table~\ref{exp:comparison_on_dialoguereasoningplanning}. \modelname{} significantly outperforms previous models, achieving state-of-the-art results across all evaluation metrics.

\paragraph{3D planning.}
This task aims to generate sequential action plans based on instructions and 3D contextual understanding. We evaluate on the 3D-LLM~\cite{hong2023dllm} dataset. As reported in Table~\ref{exp:comparison_on_dialoguereasoningplanning}, \modelname{} surpasses all baselines across all evaluation criteria.

\begin{table}[t!]
    \centering
    \caption{\textbf{3D visual grounding} results on ScanRefer~\cite{chen2020scanrefer} and Nr3D~\cite{achlioptas2020referit3d}.}
    \resizebox{\linewidth}{!}{
        \begin{tabular}{lcccc}
        \toprule
        \multirow{2}[2]{*}{Method} & Nr3D & & \multicolumn{2}{c}{ScanRefer}\\
        \cmidrule(lr){4-5}
        & Acc@0.25 & & Acc@0.5 & Acc@0.25 \\
        \midrule
        3DVG-Trans~\cite{zhao2021_3DVG_Transformer} & 40.80 & & 34.70 & 47.60 \\
        TGNN~\cite{huang2021text} & 37.30 & & 29.70 & 37.37 \\
        TransRefer3D~\cite{transrefer3d} & 48.00 & & - & - \\
        InstanceRefer~\cite{yuan2021instancerefer} & 38.80 & & 32.93 & 40.23  \\
        FFL-3DOG~\cite{feng2021free} & 41.70 & & 34.01 & 41.33 \\
        LAR~\cite{bakr2022look} & 48.90 & & - & - \\
        SAT~\cite{yang2021sat} & 56.50 & & 30.14 & 44.54 \\
        3D-SPS~\cite{luo20223d} & 51.50 & & 36.98 & 48.82 \\
        3DJCG~\cite{cai20223djcg} & - & & 37.33 & 49.56 \\
        BUTD-DETR~\cite{jain2022bottom} & 54.60 & & 39.80 & 52.20 \\
        MVT~\cite{huang2022multi} & 59.50 & & 33.26 & 40.80 \\
        ViL3DRel~\cite{chen2022language} & 64.40 & & 37.73 & 47.94 \\
        EDA~\cite{wu2023eda} & 52.10 & & 42.26 & 54.59 \\
        3D-VisTA~\cite{zhu20233d-vista} & 64.20 & & 45.80 & 50.60 \\ 
        SceneVerse~\cite{jia2024sceneverse} & 64.90 & & 48.10 & - \\
        ChatScene~\cite{huang2024chatscene} & - & & 50.20 & 55.50 \\
        LLaVA-3D~\cite{zhu2024llava}        & - & & 42.70 & 50.10  \\
        Video-3D LLM~\cite{video3dllm} & - & & 51.72 & 58.12 \\
        GPT4Scene~\cite{GPT4Scene} & - & & 57.00 & 62.60 \\
        MiKASA~\cite{chang2024mikasa} & 64.40 & & - & - \\
        Scene-R1~\cite{scener12025} & - & & 17.10 & 38.80 \\
        \midrule
        \textbf{\modelname{} (Ours)} & \textbf{68.80} & & \textbf{59.24} & \textbf{65.85} \\
        \bottomrule
        \end{tabular}
    }
    \label{exp:comparison_on_vg}
\end{table}

\section{Limitation and Future Work}
While \modelname{} demonstrates strong reasoning performance and generalizability across diverse 3D scene understanding tasks, several limitations remain.
First, although the Scene-30K dataset provides high-quality Chain-of-Thought (CoT) supervision, it is primarily synthetic and may not fully capture the diversity and ambiguity of real-world human reasoning.
Second, the current GRPO-based RLHF optimization operates at the response level and lacks temporally grounded feedback. This limits the model’s ability to reason and act on long-range tasks in embodied settings.
Third, our dynamic view selection strategy is designed for static scenes and assumes a fixed pool of candidate views. This may restrict its applicability to real-time interactive environments.

In future work, we plan to extend \modelname{} in two key directions.
First, we will explore embodied AI in real world application that integrates path planning and action prediction with multimodal reasoning.
Second, we aim to build a world model on top of \modelname{}, enabling agents to simulate and predict future scene states.
\section{Conclusion}
In this work, we propose \modelname{}, a generalist 3D vision-language model designed to advance unified scene understanding.
To address the shortcomings of existing 3D-VLMs in reasoning generalization, we introduce Scene-30K, a large-scale, high-quality Chain-of-Thought dataset that provides structured supervision for cold start initialization. Based on this foundation, we develop a reinforcement learning framework based on Group Relative Policy Optimization (GRPO), integrating perception-based, semantics-based, and format-based rewards to refine the model’s cognitive alignment and spatial precision.
In addition, we present a dynamic view selection strategy that learns to rank multiview images based on task relevance, spatial coverage, and cross-modal alignment.
Extensive evaluations across seven representative 3D benchmarks demonstrate that \modelname{} achieves significant improvements over prior methods. Our results highlight the promise of combining structured CoT supervision, reward-driven policy optimization, and adaptive perception strategies for generalizable 3D scene understanding.

\clearpage
\appendix

\section{Ablation Study}

\paragraph{Reinforcement learning.} 
We conduct a comprehensive ablation to examine the effect of each reward function in our GRPO-based reinforcement learning.
As presented in Table~\ref{tab:ablation_rl}, reinforcement learning (RL) yields substantial improvements in both reasoning and grounding performance compared to the baseline of supervised fine-tuning (SFT). Although SFT provides strong initialization, it lacks structural regularity, spatial alignment, and semantic fidelity. The format reward enforces syntactic consistency in the output, the perception reward enhances spatial grounding through improved object localization, and the semantic reward improves alignment with the intended meaning. When combined, these reward signals lead to a significant performance increase, increasing ScanQA CIDEr from 97.95 to 106.45 and ScanRefer C@0.25 from 85.20 to 91.85. This highlights the complementary contributions of each reward component in optimizing the model’s 3D reasoning capabilities.

\begin{table}[H]
\centering
\caption{\textbf{Ablation of individual and combined rewards in GRPO-based RL.} Performance is evaluated on 3D-QA (ScanQA) and on 3D-DC (ScanRefer) tasks. And the first row corresponds to the supervised fine-tuning (SFT) baseline without any reinforcement learning.}
\label{tab:ablation_rl}
\resizebox{\linewidth}{!}{
\begin{tabular}{cccccccc}
\toprule
\multirow{2}{*}{$R_{Format}$} & \multirow{2}{*}{$R_p$} & \multirow{2}{*}{$R_{similarity}$} & \multicolumn{2}{c}{ScanQA} & & \multicolumn{2}{c}{ScanRefer}\\
                        \cline{4-5} \cline{7-8}
         &    &         & C$\uparrow$ & R$\uparrow$& & C@0.25$\uparrow$ & R@0.25$\uparrow$ \\ \midrule
  \xmark & \xmark & \xmark & 97.95  & 45.12 & & 85.20 & 55.94 \\ \midrule
  \cmark & \xmark & \xmark & 101.35 & 46.65 & & 88.00 & 57.52 \\
  \xmark & \cmark & \xmark & 102.55 & 47.34 & & 88.70 & 58.24 \\
  \xmark & \xmark & \cmark & 102.45 & 47.50 & & 88.50 & 58.33 \\ \midrule
  \cmark & \cmark & \xmark & 104.12 & 48.90 & & 89.90 & 59.75 \\
  \cmark & \xmark & \cmark & 104.75 & 49.03 & & 90.20 & 59.84 \\
  \xmark & \cmark & \cmark & 104.60 & 49.10 & & 90.10 & 59.90 \\ \midrule
  \cmark & \cmark & \cmark & \textbf{106.45} & \textbf{51.23} & & \textbf{91.85} & \textbf{62.23} \\
\bottomrule
\end{tabular}
}
\end{table}

\paragraph{Dynamic view selection.} To quantify the effect of dynamic view selection, we compare our learned strategy against three fixed-view baselines: (1) \textbf{All-views}, which uses all views of the scene; (2) \textbf{Horizontal 6-views}, comprising six front-facing views of the scene; and (3) \textbf{Bottom 6-views}, sampled from below the scene. In contrast, (4) \textbf{Ours (Learned 6-view selection)} adaptively selects the most informative six views based on learned utility scores.
As shown in Table~\ref{tab:ablation_view}, our dynamic view selection strategy consistently outperforms fixed-view baselines. On the 3D object captioning task, it improves CLIP R@1 from 30.18 with fixed horizontal 6 views to 32.23, highlighting its ability to focus on more informative visual perspectives.
Moreover, the performance gains observed on 3D visual grounding further demonstrate that adaptive view selection leads to more accurate object localization by providing contextually relevant observations.

We also study the effect of three dynamic view selection weights, which control the fusion of three scoring cues: text relevance ($w_t$), spatial coverage ($w_c$), and CLIP-based similarity ($w_{\text{clip}}$). Table~\ref{tab:ablation_view_weight} presents a grid search for various weight combinations. The results show that all three cues are complementary: using any single score alone yields suboptimal results, while balanced weighting ($w_t=0.3$, $w_c=0.5$, $w_{\text{clip}}=0.5$) achieves the best performance across tasks.

\begin{figure}[t]
    \centering
    \includegraphics[width=\linewidth]{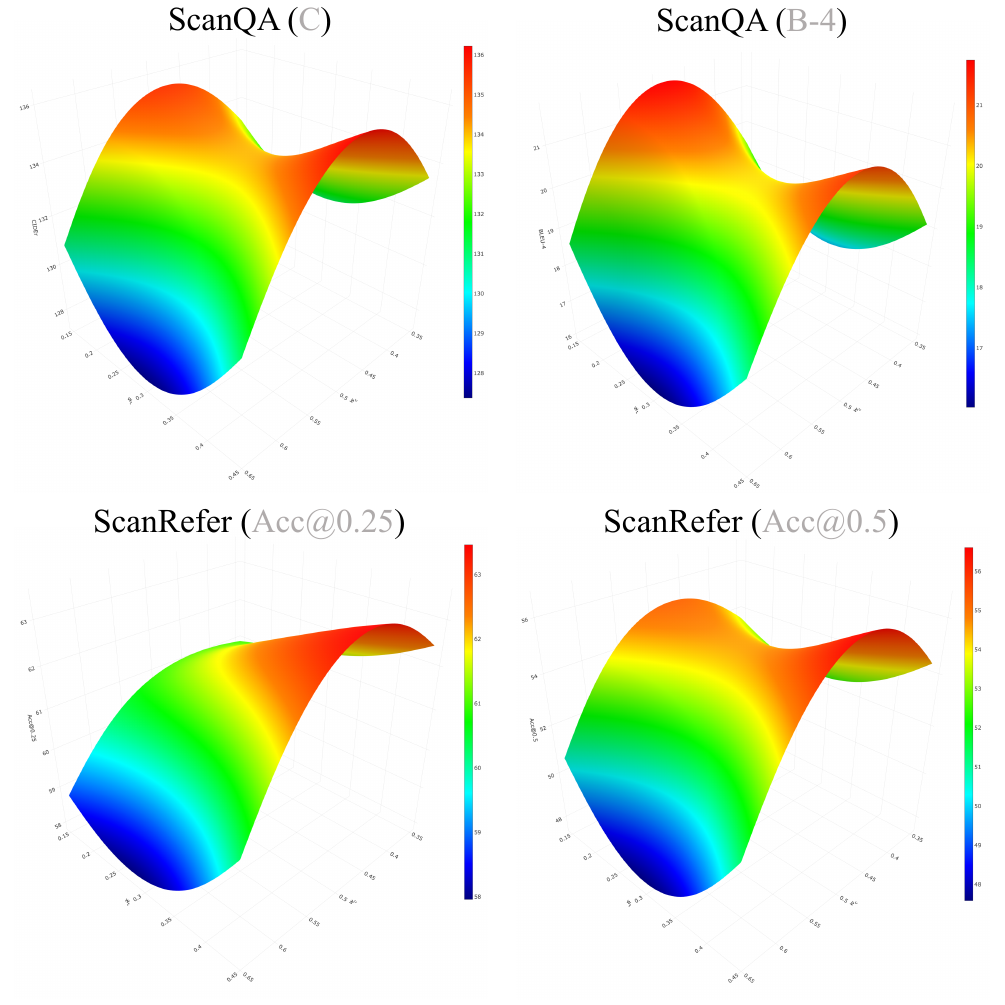}
    \caption{\textbf{Performance surfaces under different dynamic view selection weight configurations.} We analyze the influence of text relevance ($w_t$), spatial coverage ($w_c$), and CLIP-based similarity ($w_{\text{clip}}$) on model performance, with the constraint $w_c + w_{\text{clip}} = 1$. Results on 3D-QA (ScanQA) and 3D-VG (ScanRefer) reveal that optimal performance emerges when $w_t$ is within the range of 0.3 to 0.4, combined with balanced visual weights.}
    \label{fig:exp_weight}
\end{figure}

To further illustrate this, Figure~\ref{fig:exp_weight} visualizes the performance landscape over different weight configurations. The plots reveal that moderate reliance on text grounding ($w_t \approx 0.3$–$0.4$) combined with balanced visual cues leads to optimal performance, validating the effectiveness of learned weight fusion.

\paragraph{Architecture and hyperparameters.}
We conduct a step-by-step ablation to quantify the contribution of each modality encoder in our unified 3D architecture. As shown in Table~\ref{tab:ablation_arch}, we start from a baseline model using only the text and image encoder, and progressively add the depth encoder and point cloud encoder. Each modality brings clear performance gains on both 3D reasoning (SQA3D) and 3D planning (3D-LLM) tasks.
Adding the depth encoder improves performance on SQA3D, confirming that monocular geometric cues are helpful for grounding and planning. Further adding the point cloud encoder leads to larger gains, highlighting the importance of explicit 3D structure for complex reasoning. The full model (\modelname{}) achieves the best performance across all metrics.

\begin{table}[t]
\centering
\caption{\textbf{Effect of dynamic view selection.} Comparison of different view selection strategies for 3D object captioning (Cap3D) and 3D-VG (ScanRefer). Our learned selection of six optimal views achieves superior performance over fixed-view baselines.}
\label{tab:ablation_view}
\resizebox{\linewidth}{!}{
\begin{tabular}{lcccc}
\toprule
\multirow{2}{*}{View Strategy} & Cap3D & & \multicolumn{2}{c}{ScanRefer} \\
\cline{2-2}\cline{4-5}
        & CLIP R@1$\uparrow$ & & Acc@0.25$\uparrow$ & Acc@0.5$\uparrow$ \\
\midrule
All-views          & 29.19 && 61.25 & 51.73 \\
Horizontal 6-views & 30.18 && 60.53 & 50.26 \\
Bottom 6-views     & 6.63  && 57.89 & 47.63 \\
\midrule
\textbf{Learned 6-view selection (Ours)}      & \textbf{32.23} && \textbf{65.85} & \textbf{59.24} \\
\bottomrule
\end{tabular}
}
\end{table}

\begin{table}[t]
\centering
\caption{\textbf{Grid search on view weight configurations.} Performance is evaluated on 3D-QA (ScanQA) and on 3D-VG (ScanRefer) tasks.}
\label{tab:ablation_view_weight}
\resizebox{\linewidth}{!}{\begin{tabular}{ccccccccc}
\toprule
\multicolumn{3}{c}{View weight} & & \multicolumn{2}{c}{ScanQA} & & \multicolumn{2}{c}{ScanRefer} \\
\cline{1-3} \cline{5-6} \cline{8-9}
 $w_t$ & $w_c$ & $w_{\text{clip}}$ & & C$\uparrow$ & B-4$\uparrow$& & Acc@0.25 & Acc@0.5\\
 \midrule
0.3 & 0.6 & 0.4 & & 122.76 & 12.98 & & 55.34 & 42.98 \\
0.3 & 0.4 & 0.6 & & 128.49 & 15.34 & & 60.45 & 50.23 \\
0.4 & 0.5 & 0.5 & & 137.78 & 22.23 & & 63.98 & 57.95 \\
0.2 & 0.5 & 0.5 & & 136.67 & 22.80 & & 60.45 & 55.94 \\
\midrule
\textbf{0.3} & \textbf{0.5} & \textbf{0.5} & & \textbf{138.67} & \textbf{23.56} & & \textbf{65.85} & \textbf{59.24} \\
\bottomrule
\end{tabular}
}
\end{table}

\begin{table}[t]
\centering
\caption{\textbf{Incremental modality encoder ablation starting from Text \& Image encoder.} Performance is evaluated on 3D reasoning (SQA3D) and 3D planning (3D-LLM) tasks. The first row is the baseline, and each subsequent row adds one encoder. The final row (\modelname{}) includes all modalities.}
\label{tab:ablation_arch}
\resizebox{\linewidth}{!}{
\begin{tabular}{lccccc}
\toprule
\multirow{2}{*}{Setting} & \multicolumn{2}{c}{SQA3D} & & \multicolumn{2}{c}{3D-LLM} \\
\cline{2-3} \cline{5-6}
        & C$\uparrow$ & B-4$\uparrow$ & & C$\uparrow$ & B-4$\uparrow$ \\
\midrule
Text \& Image Encoder & 110.23 & 15.34 & & 200.45 & 20.15 \\
+ Depth Encoder & 115.23 & 18.34 & & 205.45 & 21.15 \\
+ Point Encoder & 120.12 & 20.13 & & 215.34 & 22.34 \\ \midrule
 \textbf{\modelname{} (Ours)}  & \textbf{138.67} & \textbf{23.56} & & \textbf{230.50} & \textbf{25.45} \\
\bottomrule
\end{tabular}
}
\end{table}

\begin{table}[t]
\centering
\caption{\textbf{Ablation of LoRA rank $\delta$.} Increasing rank improves performance up to a point, with diminishing returns beyond $\delta=12$. Performance is evaluated on 3D-QA (ScanQA) and on 3D-VG (Nr3D) tasks.}
\label{tab:ablation_lora_rank}
\resizebox{\linewidth}{!}{\begin{tabular}{cccccccc}
\toprule
\multirow{2}{*}{LoRA Rank $\delta$} & \multirow{2}{*}{Params (M)} & \multicolumn{4}{c}{ScanQA} & & Nr3D \\
\cline{3-6} \cline{8-8}
   & & C$\uparrow$ & B-4$\uparrow$ & M$\uparrow$ & R$\uparrow$  && Acc@0.25\\
\midrule
4         & 82  & 94.57  & 13.34 & 17.12 & 47.23 && 63.12 \\
8         & 112 & 101.69 & 15.34 & 20.12 & 49.23 && 65.43 \\ \midrule
\textbf{12 (Ours)} & 142 & \textbf{106.45} & \textbf{17.80} & \textbf{22.13} & \textbf{51.23} && \textbf{68.80}  \\ \midrule
16        & 175 & 106.79 & 17.45 & 22.23 & 51.33 && 68.82 \\
32        & 250 & 107.01 & 17.90 & 22.50 & 51.45 && 68.90 \\
\bottomrule
\end{tabular}
}
\end{table}

Finally, we examine the impact of the LoRA rank $\delta$, which controls the internal dimensionality of the adapter layers.
A higher rank allows for more expressive adaptation but increases the number of trainable parameters.
As shown in Table~\ref{tab:ablation_lora_rank}, increasing $\delta$ from 4 to 12 results in significant performance gains across reasoning and grounding tasks, with ScanQA CIDEr improving from 94.57 to 106.45, and Nr3D accuracy rising from 63.12 to 68.80.
However, the performance gains begin to saturate beyond $\delta=12$, as further increasing the rank to 32 yields only marginal improvements at the cost of higher parameter overhead. These results suggest that $\delta=12$ offers the best trade-off between performance and efficiency.

\section{Implementations Details}
\begin{figure*}[t]
\centering
\small %
\resizebox{\textwidth}{!}{ %
\begin{minipage}{\textwidth}
\begin{tcolorbox}[colback=cyan!5, colframe=gray!30, left=6pt, right=6pt, top=6pt, bottom=6pt, title={\Large\textbf{Prompt}}, fonttitle=\bfseries, coltitle=black, colbacktitle=cyan!5, center title, sharp corners=south,boxrule=0.4pt,rounded corners, after title={}]

You are an AI visual assistant in a 3D scene. Each scene contains a piece of description as follows. \\[2pt]
Scene description of the scene: In this apartment scene, there is a floor, sink, mirror, desk, clock, scale, kitchen cabinets, cabinets, tables, toaster, stools, bed, trash cans, dish rack, curtains, tissue box, toilet, bicycle, shelf, and a guitar case.
The sink is in front of the guitar case, while the cabinet is to the left of the guitar case.
The stool is in front of the scale and the toilet. The bicycle is adjacent to the cabinet, and the shelf is in front of the sink.
The floor supports the stool and the sink. The scene suggests a functional and organized apartment with various objects for daily activities.\\[2pt]
\textbf{Following the scene description, think step by step and answer the following question directly with following output format}. \\
\textbf{\textcolor{red}{Output format:}} \textcolor[RGB]{142,217,115}{\textless think\textgreater}...\ reasoning...\textcolor[RGB]{142,217,115}{\textless/think\textgreater} \textcolor[RGB]{142,217,115}{\textless answer\textgreater}...\ final answer...\textcolor[RGB]{142,217,115}{\textless/answer\textgreater}. \\[6pt]
\textbf{Question:} \colorbox{orange!20}{\textcolor{black}{What is in the right corner of room by curtains?}}
\begin{center}
    \includegraphics[width=\linewidth]{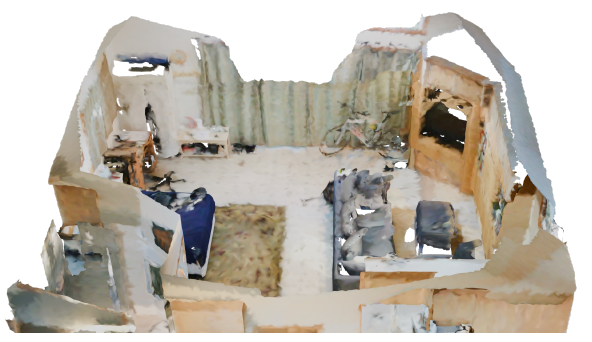}
\end{center}
\end{tcolorbox}

\begin{tcolorbox}[colback={green!5!white}, colframe=gray!30, left=6pt, right=6pt, top=6pt, bottom=6pt, title={\Large\textbf{Response}}, fonttitle=\bfseries, coltitle=black, colbacktitle={green!5!white}, center title, sharp corners=south, boxrule=0.4pt,rounded corners,after title={}]

\textcolor[RGB]{142,217,115}{\textless think\textgreater} \\[2pt]
\textbf{Step 1:} Locate the spatial reference \\ 
\quad \textbf{``\textit{right corner by curtains}''} The question specifies the right corner of the room and objects near the curtains.\\
This sets our spatial query target. \\[2pt]
\textbf{Step 2:} Identify all mentions of curtains and associated objects \\
\quad A green curtain is described as behind a bicycle.\\
\quad A tan curtain is above a desk near the bed.\\
Of these, the green curtain is more likely to be in a corner, since the desk bed setup \textcolor{blue!80!black}{implies a central or side-wall arrangement.}\\[2pt]
\textbf{Step 3:} Explore the green curtain region \\
Near the green curtain: a bicycle and a cabinet. The cabinet is described as housing a TV, and placed opposite the kitchen and sofa. \\
\textcolor{blue!80!black}{This implies a corner-located entertainment area.} \\[2pt]
\textbf{Step 4:} Evaluate dominance of objects in that corner \\
\quad The bike is adjacent, but the TV cabinet is structurally larger, described as a focal point, and clearly fixed in place. \\
Curtains likely frame the cabinet setup, \textcolor{blue!80!black}{reinforcing its positional centrality in that corner.} \\[2pt]
\textbf{Conclusion:} The \textbf{TV cabinet} is the primary object in the right corner near the \textbf{green curtains}. \\[2pt]
\textcolor[RGB]{142,217,115}{\textless/think\textgreater} \\[2pt]
\textcolor[RGB]{142,217,115}{\textless answer\textgreater} \\[2pt]
\textcolor{red!70!black}{The object in the right corner of the room by the curtains is the \textbf{TV cabinet}, which contains the TV.} \\[2pt]
\textcolor[RGB]{142,217,115}{\textless/answer\textgreater}
\end{tcolorbox}
\end{minipage}
}
\caption{\textbf{Scene-30K CoT data example 1.}}
\label{fig:cot_example_1}
\end{figure*}

\begin{figure*}[t]
\centering
\small %
\resizebox{\textwidth}{!}{ %
\begin{minipage}{\textwidth}
\begin{tcolorbox}[colback=cyan!5, colframe=gray!30, left=6pt, right=6pt, top=6pt, bottom=6pt, title={\Large\textbf{Prompt}}, fonttitle=\bfseries, coltitle=black, colbacktitle=cyan!5, center title, sharp corners=south,boxrule=0.4pt,rounded corners, after title={}]

You are an AI visual assistant in a 3D scene. Each scene contains a piece of description as follows. \\[2pt]
Scene description of the scene: In the kitchen, there is a sink, an armchair, a kitchen counter, two windows, two trash cans, two refrigerators, a floor, and ten chairs.
The chairs are positioned in front of each other and behind the trash cans. The trash cans are in front of the chairs, and the chairs are to the right of the armchair and another chair. The chairs are supported by the floor.
The kitchen appears to be a functional space with various objects for cooking, sitting, and disposing of waste.\\[2pt]
\textbf{Following the scene description, think step by step and answer the following question directly with following output format}. \\
\textbf{\textcolor{red}{Output format:}} \textcolor[RGB]{142,217,115}{\textless think\textgreater}...\ reasoning...\textcolor[RGB]{142,217,115}{\textless/think\textgreater} \textcolor[RGB]{142,217,115}{\textless answer\textgreater}...\ final answer...\textcolor[RGB]{142,217,115}{\textless/answer\textgreater}. \\[6pt]
\textbf{Question:} \colorbox{orange!20}{\textcolor{black}{What color is the refrigerator?}}
\begin{center}
\includegraphics[width=\linewidth]{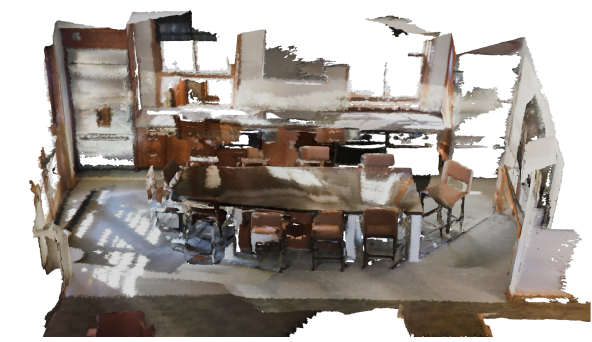}
\end{center}
\end{tcolorbox}

\begin{tcolorbox}[colback={green!5!white}, colframe=gray!30, left=6pt, right=6pt, top=6pt, bottom=6pt, title={\Large\textbf{Response}}, fonttitle=\bfseries, coltitle=black, colbacktitle={green!5!white}, center title, sharp corners=south, boxrule=0.4pt,rounded corners,after title={}]

\textcolor[RGB]{142,217,115}{\textless think\textgreater} \\[2pt]
\textbf{Step 1:} Identify the query and object \\ 
\quad The question is: \textbf{``\textit{What color is the refrigerator?}''} This requires identifying all refrigerators in the scene and resolving their associated color attributes.\\
This sets our spatial query target. \\[2pt]
\textbf{Step 2:} Collect all references to refrigerators and colors \\
\quad A stainless steel refrigerator is described with a bottom freezer. \\
\quad A white refrigerator is also mentioned. \\
This \textcolor{blue!80!black}{suggests that the scene may contain more than one refrigerator or that there is ambiguity in how a single unit is described.} \\[2pt]
\textbf{Step 3:} Determine the number of refrigerators \\
One caption explicitly states ``two refrigerators'', which strongly implies the existence of two separate appliances. \\
\textcolor{blue!80!black}{Thus, the color differences likely correspond to two distinct refrigerators rather than conflicting descriptions of one.} \\[2pt]
\textbf{Step 4:} Evaluate user intent and precision of the query \\
\quad The question is phrased as ``the refrigerator'', suggesting singular intent. \\
However, since the scene contains two refrigerators, both should be acknowledged unless additional disambiguation is given.\\[2pt]
\textbf{Conclusion:} The kitchen contains \textbf{two refrigerators}: one is \textbf{stainless steel}, and the other is \textbf{white}.\\[2pt]
\textcolor[RGB]{142,217,115}{\textless/think\textgreater} \\[2pt]
\textcolor[RGB]{142,217,115}{\textless answer\textgreater} \\[2pt]
\textcolor{red!70!black}{The \textbf{refrigerators} in the kitchen are of different colors. One is \textbf{stainless steel}, and the other is \textbf{white}.} \\[2pt]
\textcolor[RGB]{142,217,115}{\textless/answer\textgreater}
\end{tcolorbox}
\end{minipage}
}
\caption{\textbf{Scene-30K CoT data example 2.}}
\label{fig:cot_example_2}
\end{figure*}

\begin{figure*}[t]
\centering
\small %
\resizebox{\textwidth}{!}{ %
\begin{minipage}{\textwidth}
\begin{tcolorbox}[colback=cyan!5, colframe=gray!30, left=6pt, right=6pt, top=6pt, bottom=6pt, title={\Large\textbf{Prompt}}, fonttitle=\bfseries, coltitle=black, colbacktitle=cyan!5, center title, sharp corners=south,boxrule=0.4pt,rounded corners, after title={}]

You are an AI visual assistant in a 3D scene. Each scene contains a piece of description as follows. \\[2pt]
Scene description of the scene: In this classroom scene, there is a single floor, three windowsills, 22 chairs, two windows, and seven tables.
The chairs are adjacent to the tables, with some chairs behind the tables and others to the left.
The tables are positioned in front of some chairs and behind others. The tables are also stacked, with one table behind another. The chairs are resting on the floor, which supports both the chairs and the tables.
Inside one of the tables, there is a chair. The classroom appears to be well-equipped with furniture for students to sit and work comfortably.\\[2pt]
\textbf{Following the scene description, think step by step and answer the following question directly with following output format}. \\
\textbf{\textcolor{red}{Output format:}} \textcolor[RGB]{142,217,115}{\textless think\textgreater}...\ reasoning...\textcolor[RGB]{142,217,115}{\textless/think\textgreater} \textcolor[RGB]{142,217,115}{\textless answer\textgreater}...\ final answer...\textcolor[RGB]{142,217,115}{\textless/answer\textgreater}. \\[6pt]
\textbf{Question:} \colorbox{orange!20}{\textcolor{black}{Please describe the table.}}
\begin{center}
\includegraphics[width=\linewidth]{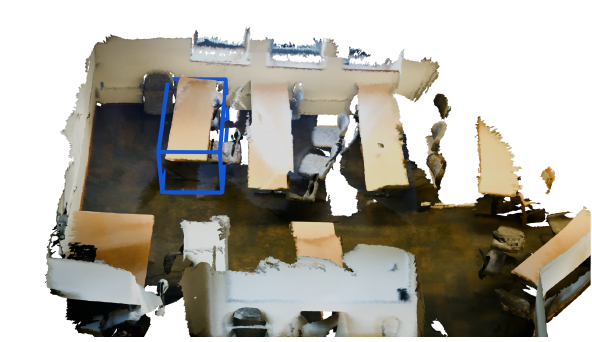}
\end{center}
\end{tcolorbox}

\begin{tcolorbox}[colback={green!5!white}, colframe=gray!30, left=6pt, right=6pt, top=6pt, bottom=6pt, title={\Large\textbf{Response}}, fonttitle=\bfseries, coltitle=black, colbacktitle={green!5!white}, center title, sharp corners=south, boxrule=0.4pt,rounded corners,after title={}]

\textcolor[RGB]{142,217,115}{\textless think\textgreater} \\[2pt]
\textbf{Step 1:} Localize the queried table \\ 
\quad The question refers to \textbf{``\textit{the table}''}, and the blue 3D bounding box highlights a specific rectangular table placed in the classroom scene.\\
\quad Given the presence of seven tables, this selection \textcolor{blue!80!black}{implies that the referred table is either spatially unique, visually distinct, or structurally notable}. \\[2pt]
\textbf{Step 2:} Analyze spatial context and adjacency \\
\quad This table is positioned \textbf{near the wall and windows}, \textcolor{blue!80!black}{suggesting it is in a front or peripheral position relative to the rest of the layout}.\\[2pt]
\textbf{Step 3:} Interpret table arrangement and visibility \\
\quad Its orientation is consistent with the row-style classroom layout. The surface color is darker than adjacent tables, possibly due to lighting, material, or rendering artifacts, which may \textcolor{blue!80!black}{imply a distinct role or usage}. \\[2pt]
\textbf{Step 4:} Evaluate its role and prominence \\
\quad Being placed near the windows and slightly apart from the central clutter of chairs \textcolor{blue!80!black}{suggests this table may be used by an instructor or for special functions}.\\[2pt]
\textbf{Conclusion:} The table highlighted in blue box is a \textbf{rectangular classroom table}, located near the wall and windows. It likely serves an active role in the classroom, possibly for instruction or individual work.\\[2pt]
\textcolor[RGB]{142,217,115}{\textless/think\textgreater} \\[2pt]
\textcolor[RGB]{142,217,115}{\textless answer\textgreater} \\[2pt]
\textcolor{red!70!black}{The table is a \textbf{rectangular classroom table}, located near the wall and windows.} \\[2pt]
\textcolor[RGB]{142,217,115}{\textless/answer\textgreater}
\end{tcolorbox}
\end{minipage}
}
\caption{\textbf{Scene-30K CoT data example 3.}}
\label{fig:cot_example_3}
\end{figure*}

\begin{figure*}[t]
\centering
\small %
\resizebox{\textwidth}{!}{ %
\begin{minipage}{\textwidth}
\begin{tcolorbox}[colback=cyan!5, colframe=gray!30, left=6pt, right=6pt, top=6pt, bottom=6pt, title={\Large\textbf{Prompt}}, fonttitle=\bfseries, coltitle=black, colbacktitle=cyan!5, center title, sharp corners=south,boxrule=0.4pt,rounded corners, after title={}]

You are an AI visual assistant in a 3D scene. Each scene contains a piece of description as follows. \\[2pt]
Scene description of the scene: In this office scene, there are two windows, one pair of headphones, two file cabinets, one telephone, and five monitors.
The monitors are positioned adjacent to each other and beside another monitor. Interestingly, the headphones are found inside one of the monitors. The office appears to be well-equipped with technology, suggesting a modern and efficient workspace.
The presence of file cabinets implies a need for organization and storage. Overall, this office scene exudes a professional and productive atmosphere. \\[2pt]
\textbf{Following the scene description, think step by step and answer the following question directly with following output format}. \\
\textbf{\textcolor{red}{Output format:}} \textcolor[RGB]{142,217,115}{\textless think\textgreater}...\ reasoning...\textcolor[RGB]{142,217,115}{\textless/think\textgreater} \textcolor[RGB]{142,217,115}{\textless answer\textgreater}...\ final answer...\textcolor[RGB]{142,217,115}{\textless/answer\textgreater}. \\[6pt]
\textbf{Question:} \colorbox{orange!20}{\textcolor{black}{What color does the sofa look?}}
\begin{center}
\includegraphics[width=\linewidth]{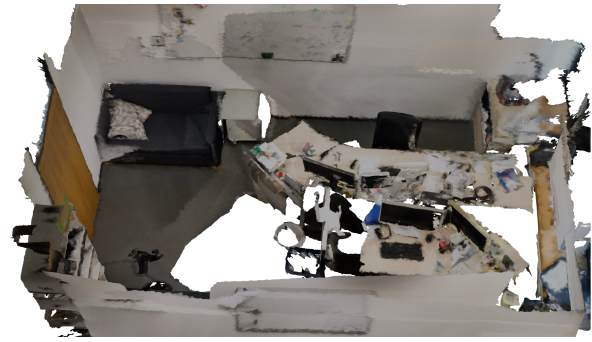}
\end{center}
\end{tcolorbox}

\begin{tcolorbox}[colback={green!5!white}, colframe=gray!30, left=6pt, right=6pt, top=6pt, bottom=6pt, title={\Large\textbf{Response}}, fonttitle=\bfseries, coltitle=black, colbacktitle={green!5!white}, center title, sharp corners=south, boxrule=0.4pt,rounded corners,after title={}]

\textcolor[RGB]{142,217,115}{\textless think\textgreater} \\[2pt]
\textbf{Step 1:} Localize the queried table \\ 
\quad The question asks: \textbf{``\textit{What color does the sofa look?}''} \quad This \textcolor{blue!80!black}{implies that we need to locate all references to the sofa in the scene and determine the most consistently described color}.\\[2pt]
\textbf{Step 2:} Extract sofa-related descriptions from captions \\
\quad The sofa is referred to as a ``\textit{dark colored two-seater futon.}''. \quad Another line mentions a``\textit{black sofa.}'' \\
These two phrases strongly \textcolor{blue!80!black}{suggest that the sofa is likely dark, with black being a probable color.} \\[2pt]
\textbf{Step 3:} Examine object-level annotations \\
Looking into the object annotations \\
\quad One mentions a ``\textit{dark colored two-seater futon located by the door,}'' \quad Another describes a ``\textit{black sofa located to the wall behind the fan.}'' \\
\quad These consistently describe the sofa using the terms black, \textcolor{blue!80!black}{which suggests that the sofa appears consistently dark in the scene.} \\[2pt]
\textbf{Step 4:} Check for contradictory evidence \\
\quad To ensure reliability, we examine whether any alternative colors are associated with the sofa. No such references are found, \textcolor{blue!80!black}{which strengthens the hypothesis that black is the intended color.} \\[2pt]
\textbf{Conclusion:} The \textbf{sofa} is most likely \textbf{black}. \\[2pt]
\textcolor[RGB]{142,217,115}{\textless/think\textgreater} \\[2pt]
\textcolor[RGB]{142,217,115}{\textless answer\textgreater} \\[2pt]
\textcolor{red!70!black}{\textbf{The sofa} in the scene is \textbf{black}.} \\[2pt]
\textcolor[RGB]{142,217,115}{\textless/answer\textgreater}
\end{tcolorbox}
\end{minipage}
}
\caption{\textbf{Scene-30K CoT data example 4.}}
\label{fig:cot_example_4}
\end{figure*}

\begin{figure*}[t]
\centering
\small %
\resizebox{\textwidth}{!}{ %
\begin{minipage}{\textwidth}
\begin{tcolorbox}[colback=cyan!5, colframe=gray!30, left=6pt, right=6pt, top=6pt, bottom=6pt, title={\Large\textbf{Prompt}}, fonttitle=\bfseries, coltitle=black, colbacktitle=cyan!5, center title, sharp corners=south,boxrule=0.4pt,rounded corners, after title={}]

You are an AI visual assistant in a 3D scene. Each scene contains a piece of description as follows. \\[2pt]
Scene description of the scene: In the Conference Room, there are two cabinets and twenty-five chairs.
The cabinets are positioned to the left and behind the chairs, while the chairs are to the right of the cabinets. Additionally, one chair is in front of the cabinet, and another chair is to the left of it.
The room appears to be a functional space for meetings or discussions, with the cabinets potentially used for storage and the chairs for seating. The arrangement suggests a professional and organized atmosphere.\\[2pt]
\textbf{Following the scene description, think step by step and answer the following question directly with following output format}. \\
\textbf{\textcolor{red}{Output format:}} \textcolor[RGB]{142,217,115}{\textless think\textgreater}...\ reasoning...\textcolor[RGB]{142,217,115}{\textless/think\textgreater} \textcolor[RGB]{142,217,115}{\textless answer\textgreater}...\ final answer...\textcolor[RGB]{142,217,115}{\textless/answer\textgreater}. \\[6pt]
\textbf{Question:} \colorbox{orange!20}{\textcolor{black}{What is the object on the wall north of the yellow table?}}
\begin{center}
\includegraphics[width=\linewidth]{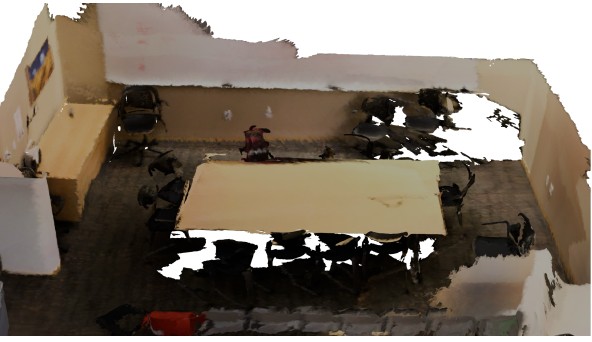}
\end{center}
\end{tcolorbox}

\begin{tcolorbox}[colback={green!5!white}, colframe=gray!30, left=6pt, right=6pt, top=6pt, bottom=6pt, title={\Large\textbf{Response}}, fonttitle=\bfseries, coltitle=black, colbacktitle={green!5!white}, center title, sharp corners=south, boxrule=0.4pt,rounded corners,after title={}]

\textcolor[RGB]{142,217,115}{\textless think\textgreater} \\[2pt]
\textbf{Step 1:} Identify the query and spatial reference \\ 
\quad The question asks: \textbf{``\textit{What is the object on the wall north of the yellow table?}''} \quad This \textcolor{blue!80!black}{implies a need to first determine the location of the yellow table and then identify the object directly north of it}.\\[2pt]
\textbf{Step 2:} Understand the scene layout \\
\quad According to the scene description, the setting is a conference room, which contains: a yellow table, chairs, cabinets, and a whiteboard. \quad The mention of a \textbf{whiteboard} in the scene immediately \textcolor{blue!80!black}{suggests it could be a candidate for a wall-mounted item in a directional context}. \\[2pt]
\textbf{Step 3:} Cross-reference with caption information \\
Scene captions state that \textit{``the whiteboard is located next to the large table''}, and in one instance, it is said to be \textit{``on the wall north of the yellow table.''} \quad This direct mention \textcolor{blue!80!black}{supports the hypothesis that the whiteboard lies on the north wall, aligning with the spatial query.} \\[2pt]
\textbf{Step 4:} Check for competing candidates \\
There are mentions of \textit{cabinets} and \textit{pictures}, but their described positions place them near corners or above other objects, not clearly on the north wall relative to the yellow table. \quad This \textcolor{blue!80!black}{strongly implies that the whiteboard is the only clearly described object positioned on the wall north of the yellow table.} \\[2pt]
\textbf{Conclusion:} The object on the wall north of \textbf{the yellow table} is most likely \textbf{a whiteboard}. \\[2pt]
\textcolor[RGB]{142,217,115}{\textless/think\textgreater} \\[2pt]
\textcolor[RGB]{142,217,115}{\textless answer\textgreater} \\[2pt]
\textcolor{red!70!black}{The object \textbf{on the wall north of the yellow table} is \textbf{a whiteboard}.} \\[2pt]
\textcolor[RGB]{142,217,115}{\textless/answer\textgreater}
\end{tcolorbox}
\end{minipage}
}
\caption{\textbf{Scene-30K CoT data example 5.}}
\label{fig:cot_example_5}
\end{figure*}

\paragraph{Data synthesis.} First, a Scene-30K dataset is synthesized using Gemini‑Pro~\cite{geminipro2025}, producing 35,248 raw CoT reasoning examples. To ensure that only high‑quality chains of thought (CoT) are retained, we design a rule‑based filtering that reduces the corpus to 30,012 examples. Some examples are visualized in Figure~\ref{fig:cot_example_1}-\ref{fig:cot_example_5}.

Specifically, the rule‑based filtering process is as follows:
We first verify that each example follows the required output format: \texttt{<think>reasoning</think><answer>final answer</answer>}. The \texttt{<think>} segment must contain at least 30 words, and the \texttt{<answer>} segment at least 20 words, to filter out overly brief reasoning and answers.
Subsequently, we assess whether the \texttt{<think></think>} segment exhibits genuine multi-step reasoning, as opposed to a single-step deduction. To ensure this, we mandate the presence of at least three explicit reasoning steps, identified through lexical cues such as ``Step n'', ``First'', ``Next'' or ``Last''. Moreover, the final step must explicitly reference the target entity posed in the question (\textit{e.g.}, ``Conclusion: ...''), as exemplified in Figure~\ref{fig:cot_example_1}–\ref{fig:cot_example_5}.
Finally, we assess the logical consistency between the reasoning and the answer. Specifically, we prompt Gemini 2.5 Pro~\cite{geminipro2025} with the pair \{\textit{think}, question\}, where \textit{think} refers to the reasoning content enclosed within the \texttt{<think></think>} tags. The model is asked to independently generate an answer $\hat{a}$.
A sample is retained only if the normalized Levenshtein similarity between $\hat{a}$ and the content within the \texttt{<answer></answer>} tags, denoted as $a$, is at least 0.8. The similarity score is defined as:
\begin{equation}
    \text{Similarity} (\hat{a},a )=1-\frac{D_{\text{lev}}(\hat{a},a)}{\text{max}(\vert \hat{a}\vert ,\vert a \vert  ) } ,
\end{equation}
where $D_{\text{lev}}(\hat{a}, a)$ denotes the Levenshtein distance, and $|\cdot|$ represents the character length of the string.
If the score falls below 0.8, the sample is discarded, even if it satisfies the format and step-count criteria. 

The complete filtering procedure is summarized in Algorithm~\ref{alg:filter}. After applying all filtering criteria, Scene-30K dataset is constituted and serves as the cold-start initialization for \modelname{}.
\begin{algorithm}[t]
    \caption{Rule-based Filtering for Scene-30K}
    \label{alg:filter}
    \begin{algorithmic}[1]
        \REQUIRE Raw CoT examples $\mathcal{D}_{\text{raw}} = \{(q_i, t_i, a_i)\}_{i=1}^{N}$
        \ENSURE Filtered CoT dataset $\mathcal{D}_{\text{final}}$
        \STATE $\mathcal{D}_{\text{final}} \gets \emptyset$
        \FORALL{ $(q, t, a)$ in $\mathcal{D}_{\text{raw}}$}
            \IF{format is invalid}
                \STATE \textbf{continue}
            \ENDIF
            \IF{word count of $t <30$ or word count of $a < 20$}
                \STATE \textbf{continue}
            \ENDIF
            \IF{number of reasoning steps in $t<3$}
                \STATE \textbf{continue}
            \ENDIF
            \IF{final step in $t$ does not mention target entity}
                \STATE \textbf{continue}
            \ENDIF
            \STATE Prompt Gemini 2.5 Pro with $(t, q)$ to generate predicted answer $\hat{a}$
            \STATE Compute Levenshtein similarity score: $s = 1-\frac{D_{\text{lev}}(\hat{a},a)}{\text{max}(\vert \hat{a}\vert ,\vert a \vert  ) }$
            \IF{$s < 0.8$}
                \STATE \textbf{continue}
            \ENDIF
            \STATE Add $(q, t, a)$ to $\mathcal{D}_{\text{final}}$
        \ENDFOR
        \RETURN $\mathcal{D}_{\text{final}}$
    \end{algorithmic}
\end{algorithm}

\paragraph{Architecture.} We construct the encoder and decoder components on top of the base VLM, Qwen2.5-VL-7B-Instruct\cite{Qwen2.5-VL}. We adopt SigLIP-2 (ViT-L/14)~\cite{siglip2}, Depth-Anything v2 (ViT-L/14)~\cite{yang2024depthv2}, and Point Transformer v3~\cite{wu2024ptv3} as image, depth and point cloud encoders, respectively.
The output from each encoder is linearly projected to match the dimensionality of the text tokens and concatenated with them to form a unified sequence.
And we freeze the entire backbone, including the text encoder and decoder, and fine-tune only the 12-layer LoRA adapters, the image encoder, the point cloud encoder, the depth encoder, and the dense decoder.

\paragraph{Parameter efficient tuning.} To enable efficient fine-tuning, we inject LoRA adapters~\cite{hu2022lora} into the last 8 transformer blocks of the base VLM, which comprises 28 transformer blocks. 
In each selected block, LoRA is implemented for all projection matrices in the VLM, \textit{i.e.}, $(W_q, W_k, W_v, W_o)$ in attention modules and $(W_{\text{gate}}, W_{\text{up}}, W_{\text{down}})$ in MLPs. 
Each adapter is configured with rank $\delta=12$, scaling factor $\alpha=16$, and no dropout, introducing only $\sim$12M additional trainable parameters, which account for approximately $0.17\%$ of the full backbone. 
In total, $\sim$142M parameters are updated during training, compared to $\sim$7B in full fine-tuning, resulting in a $\sim$98\% reduction in the trainable parameters. 
Only these LoRA parameters, along with the image encoder, depth encoder, point cloud encoder, and the dense decoder are updated, while all remaining backbone weights are kept frozen.

Supervised fine-tuning (SFT) is performed on Scene-30K for 2 epochs with a batch size of 12, adopting the AdamW optimizer with a weight decay of 0.1 and a cosine annealing learning rate schedule that decays from $10^{-5}$ to $10^{-6}$.
Following supervised fine-tuning (SFT), we further optimize the model via reinforcement learning  using Group Relative Policy Optimization (GRPO). The RL stage is performed for 2 epochs with a batch size of 12, employing the Adam optimizer and a fixed learning rate of $10^{-6}$. To ensure stability, a KL divergence penalty with coefficient $\beta = 0.02$ is imposed between the current policy and the frozen SFT model.

Furthermore, we introduce a dynamic view selection strategy applied during both training and inference. Given a 3D scene with a pool of multiview images, we extract visual features for each view using a pretrained SigLIP-2 encoder. For each view, we compute three complementary scores, which are aggregated using learnable weights to derive a final utility score. Following prior work~\cite{luo2024view}, we select the top-6 views ranked by this score and feed them into the model alongside corresponding depth inputs.
All experiments are conducted on 4 $\times$ NVIDIA H20 GPUs.

\section{Visualization}

\begin{figure*}
    \centering
    \includegraphics[width=\linewidth]{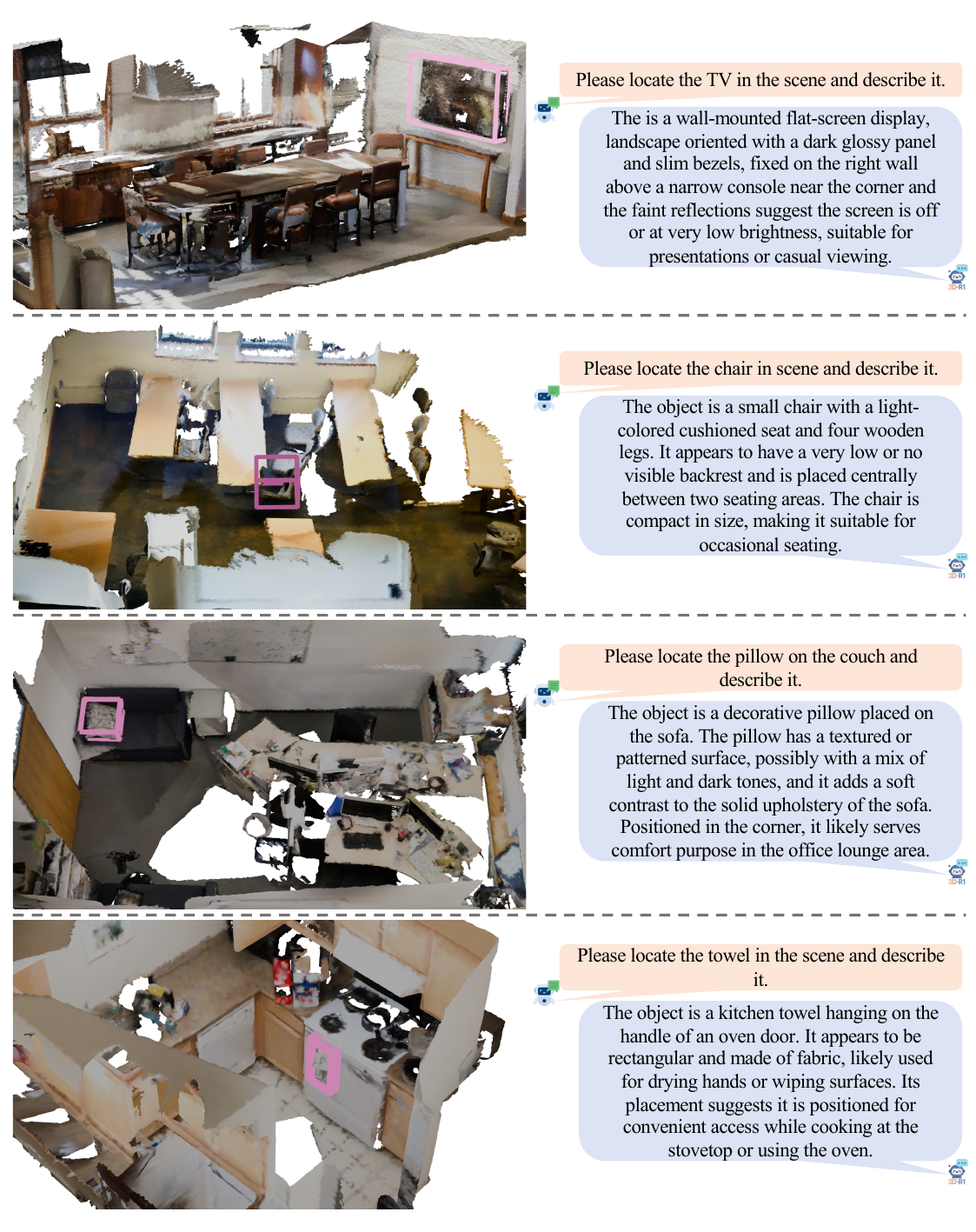}
    \caption{\textbf{Qualitative results for 3D scene dense captioning (3D-DC).}}
    \label{fig:qualitative-dc}
\end{figure*}

\begin{figure*}
    \centering
    \includegraphics[width=\linewidth]{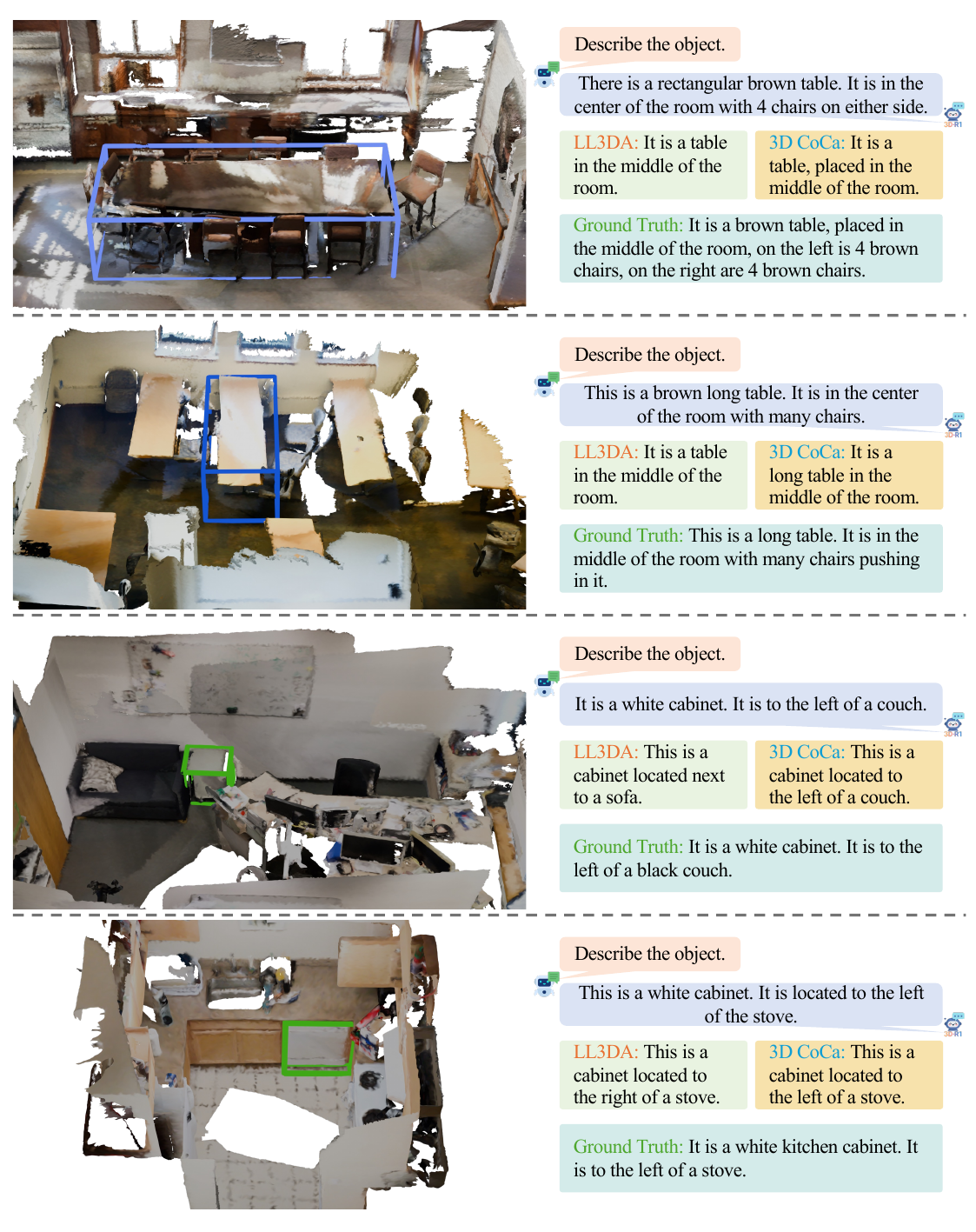}
    \caption{\textbf{Qualitative results for 3D object captioning.}}
    \label{fig:qualitative-object_captioning}
\end{figure*}

\begin{figure*}
    \centering
    \includegraphics[width=\linewidth]{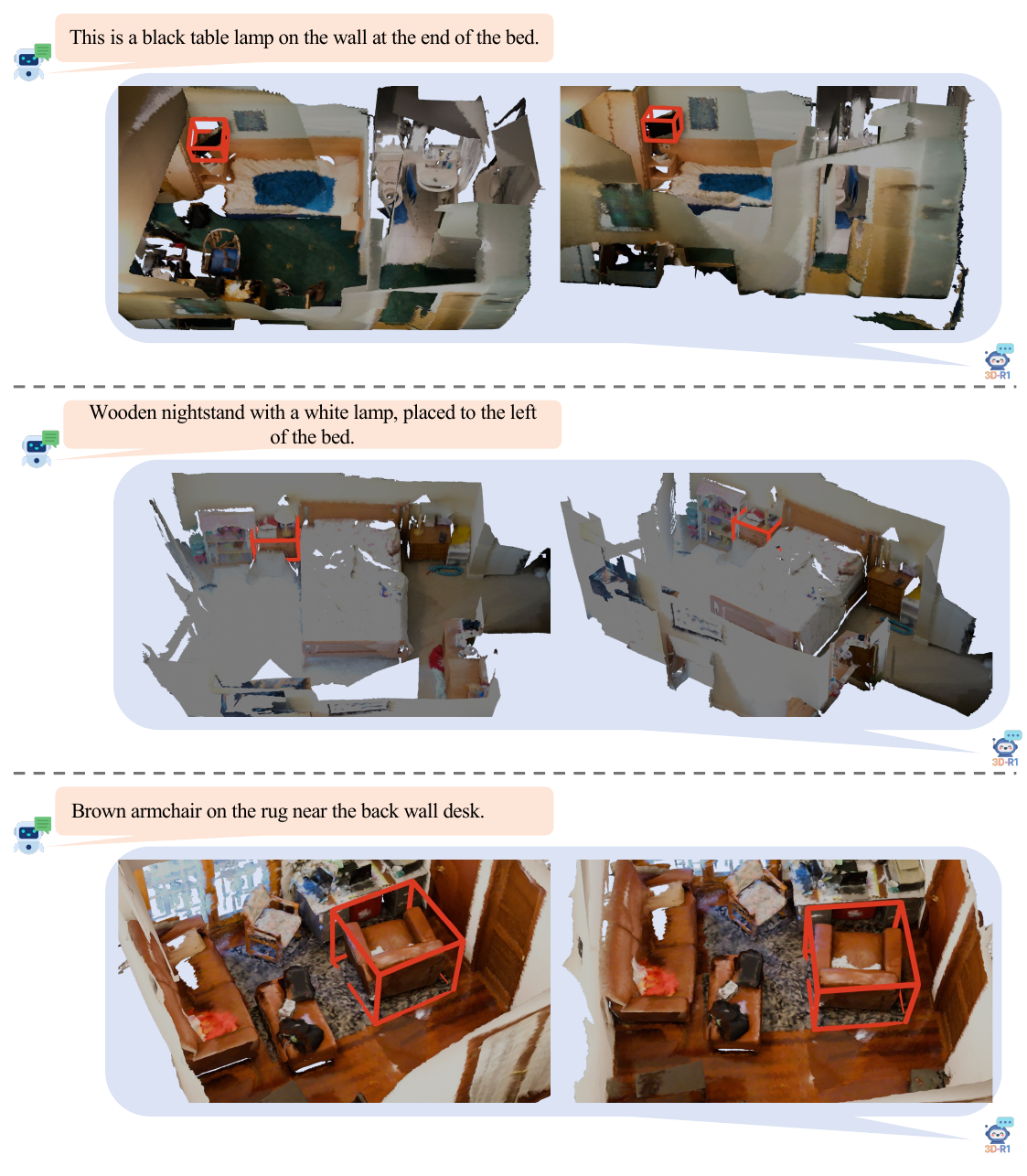}
    \caption{\textbf{Qualitative results for 3D visual grounding (3D-VG).}}
    \label{fig:qualitative-vg}
\end{figure*}

\begin{figure*}
    \centering
    \includegraphics[width=\linewidth]{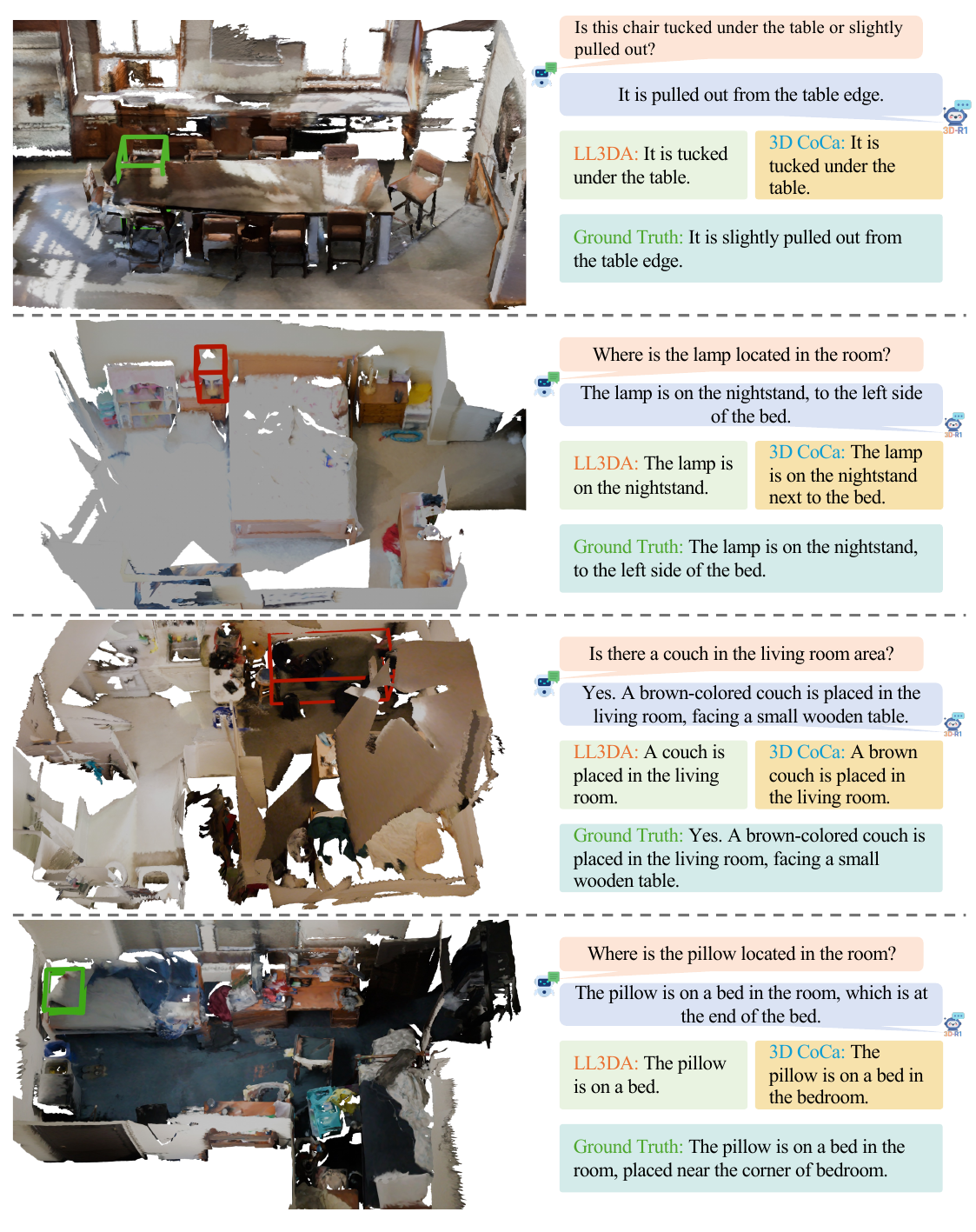}
    \caption{\textbf{Qualitative results for 3D question answering (3D-QA).}}
    \label{fig:qualitative-qa}
\end{figure*}

\begin{figure*}
    \centering
    \includegraphics[width=\linewidth]{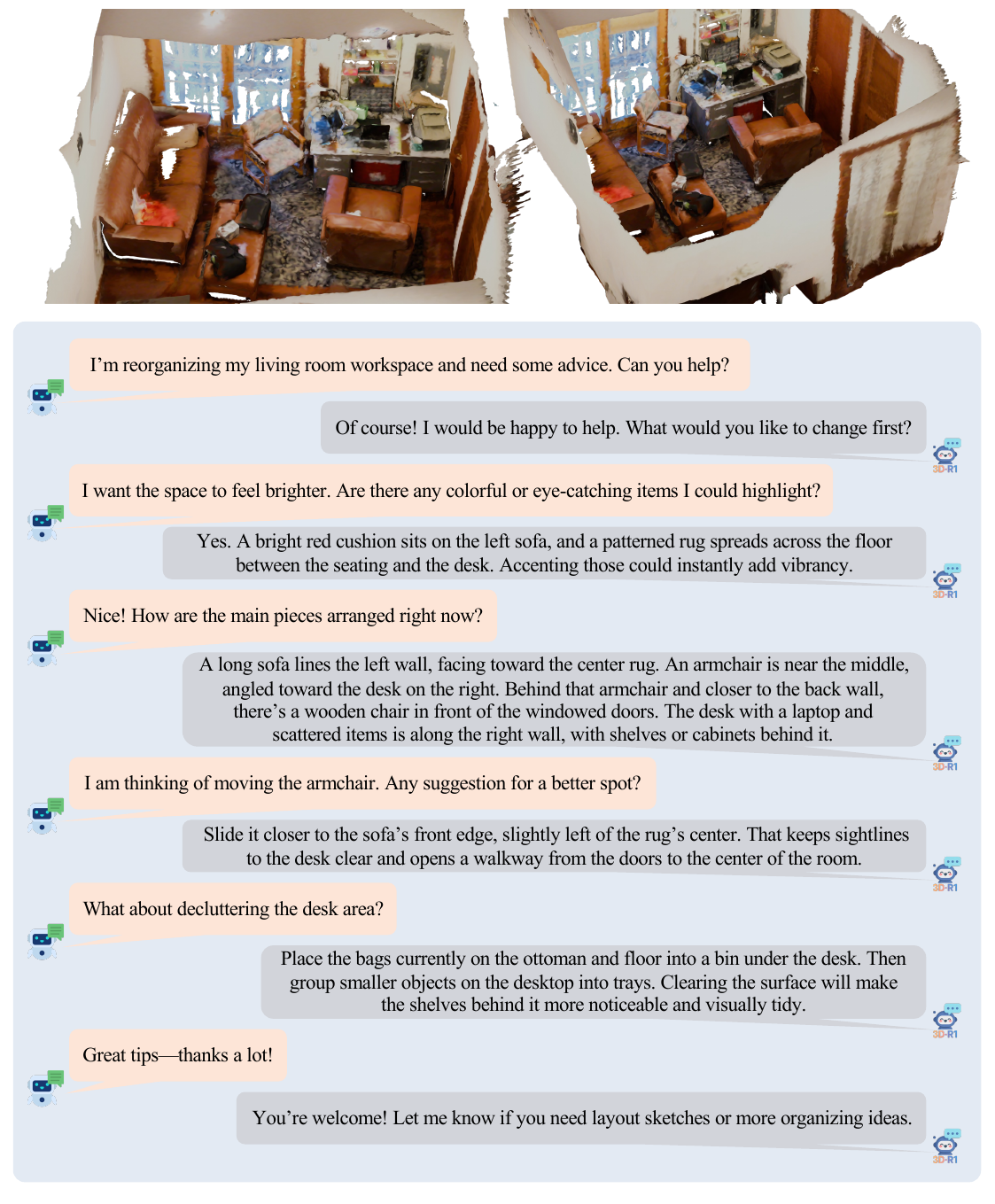}
    \caption{\textbf{Qualitative results for 3D dialogue.}}
    \label{fig:qualitative-dialogue}
\end{figure*}

\begin{figure*}
    \centering
    \includegraphics[width=\linewidth]{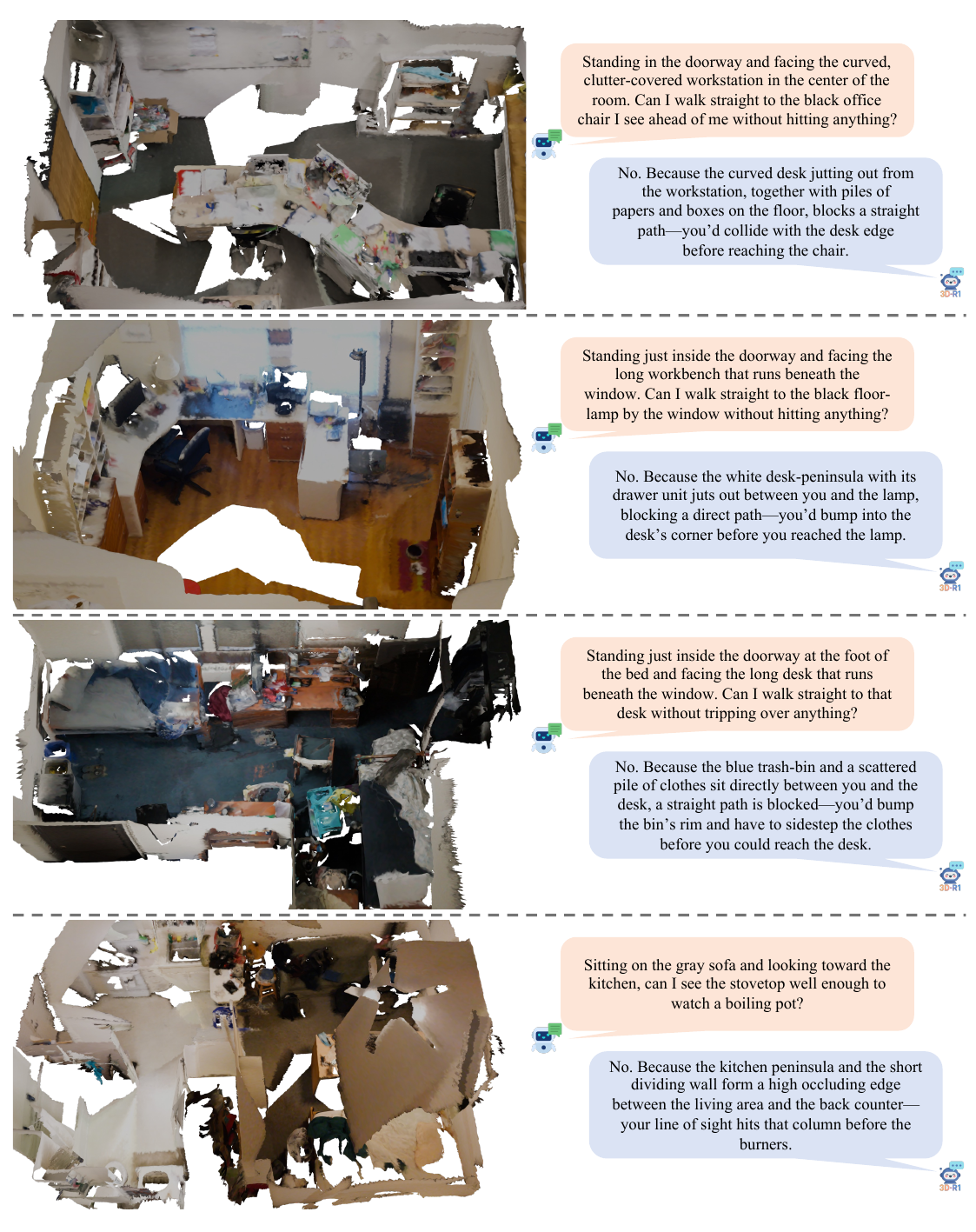}
    \caption{\textbf{Qualitative results for 3D reasoning.}}
    \label{fig:qualitative-reasoning}
\end{figure*}

\begin{figure*}
    \centering
    \includegraphics[width=\linewidth]{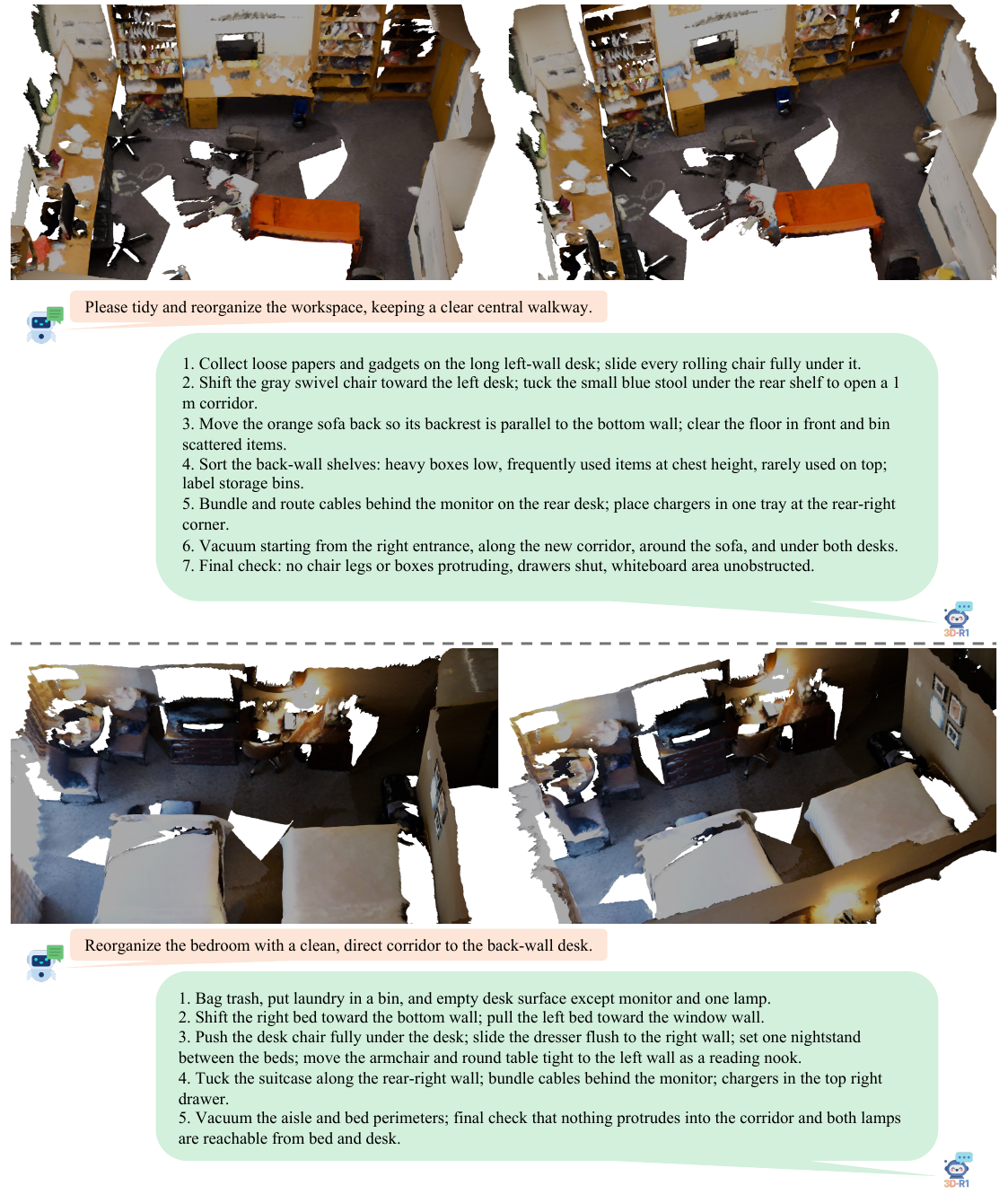}
    \caption{\textbf{Qualitative results for 3D planning.}}
    \label{fig:qualitative-planning}
\end{figure*}

To qualitatively assess the capabilities of \modelname{} in various 3D scene understanding tasks, we provide visualizations in Figures~\ref{fig:qualitative-dc}-\ref{fig:qualitative-planning}. These examples highlight the reasoning ability of the model, spatial comprehension, and multimodal alignment.


\clearpage
\clearpage
\begin{thebibliography}{71}
\providecommand{\natexlab}[1]{#1}

\bibitem[{Achlioptas et~al.(2020)Achlioptas, Abdelreheem, Xia, Elhoseiny, and Guibas}]{achlioptas2020referit3d}
Achlioptas, P.; Abdelreheem, A.; Xia, F.; Elhoseiny, M.; and Guibas, L. 2020.
\newblock Referit3d: Neural listeners for fine-grained 3d object identification in real-world scenes.
\newblock In \emph{European Conference on Computer Vision}, 422--440. Springer.

\bibitem[{Azuma et~al.(2022)Azuma, Miyanishi, Kurita, and Kawanabe}]{azuma_2022_CVPR}
Azuma, D.; Miyanishi, T.; Kurita, S.; and Kawanabe, M. 2022.
\newblock ScanQA: 3D Question Answering for Spatial Scene Understanding.
\newblock In \emph{Proceedings of the IEEE/CVF Conference on Computer Vision and Pattern Recognition (CVPR)}.

\bibitem[{Bai et~al.(2025)Bai, Chen, Liu, Wang, Ge, Song, Dang, Wang, Wang, Tang, Zhong, Zhu, Yang, Li, Wan, Wang, Ding, Fu, Xu, Ye, Zhang, Xie, Cheng, Zhang, Yang, Xu, and Lin}]{Qwen2.5-VL}
Bai, S.; Chen, K.; Liu, X.; Wang, J.; Ge, W.; Song, S.; Dang, K.; Wang, P.; Wang, S.; Tang, J.; Zhong, H.; Zhu, Y.; Yang, M.; Li, Z.; Wan, J.; Wang, P.; Ding, W.; Fu, Z.; Xu, Y.; Ye, J.; Zhang, X.; Xie, T.; Cheng, Z.; Zhang, H.; Yang, Z.; Xu, H.; and Lin, J. 2025.
\newblock Qwen2.5-VL Technical Report.
\newblock \emph{arXiv preprint arXiv:2502.13923}.

\bibitem[{BAKR, Alsaedy, and Elhoseiny(2022)}]{bakr2022look}
BAKR, E.~M.; Alsaedy, Y.~Y.; and Elhoseiny, M. 2022.
\newblock Look Around and Refer: 2D Synthetic Semantics Knowledge Distillation for 3D Visual Grounding.
\newblock In Oh, A.~H.; Agarwal, A.; Belgrave, D.; and Cho, K., eds., \emph{Advances in Neural Information Processing Systems}.

\bibitem[{Banerjee and Lavie(2005)}]{meteor2005}
Banerjee, S.; and Lavie, A. 2005.
\newblock {METEOR}: An Automatic Metric for {MT} Evaluation with Improved Correlation with Human Judgments.
\newblock In Goldstein, J.; Lavie, A.; Lin, C.-Y.; and Voss, C., eds., \emph{Proceedings of the {ACL} Workshop on Intrinsic and Extrinsic Evaluation Measures for Machine Translation and/or Summarization}, 65--72. Ann Arbor, Michigan: Association for Computational Linguistics.

\bibitem[{Cai et~al.(2022)Cai, Zhao, Zhang, Sheng, and Xu}]{cai20223djcg}
Cai, D.; Zhao, L.; Zhang, J.; Sheng, L.; and Xu, D. 2022.
\newblock 3DJCG: A Unified Framework for Joint Dense Captioning and Visual Grounding on 3D Point Clouds.
\newblock In \emph{Proceedings of the IEEE/CVF Conference on Computer Vision and Pattern Recognition}, 16464--16473.

\bibitem[{Chang et~al.(2024)Chang, Wang, Pagani, and Stricker}]{chang2024mikasa}
Chang, C.-P.; Wang, S.; Pagani, A.; and Stricker, D. 2024.
\newblock MiKASA: Multi-Key-Anchor \& Scene-Aware Transformer for 3D Visual Grounding.
\newblock In \emph{Proceedings of the IEEE/CVF Conference on Computer Vision and Pattern Recognition}, 14131--14140.

\bibitem[{Chen et~al.(2021{\natexlab{a}})Chen, Gholami, Niesner, and Chang}]{scan2cap_2021}
Chen, D.; Gholami, A.; Niesner, M.; and Chang, A. 2021{\natexlab{a}}.
\newblock Scan2Cap: Context-aware Dense Captioning in RGB-D Scans.
\newblock In \emph{2021 IEEE/CVF Conference on Computer Vision and Pattern Recognition (CVPR)}.

\bibitem[{Chen et~al.(2023{\natexlab{a}})Chen, Hu, Chen, Nießner, and Chang}]{unit3d2023}
Chen, D.; Hu, R.; Chen, X.; Nießner, M.; and Chang, A. 2023{\natexlab{a}}.
\newblock UniT3D: A Unified Transformer for 3D Dense Captioning and Visual Grounding.
\newblock In \emph{2023 IEEE/CVF International Conference on Computer Vision (ICCV)}, 18063–18073.

\bibitem[{Chen, Chang, and Nie{\ss}ner(2020)}]{chen2020scanrefer}
Chen, D.~Z.; Chang, A.~X.; and Nie{\ss}ner, M. 2020.
\newblock Scanrefer: 3d object localization in rgb-d scans using natural language.
\newblock In \emph{European Conference on Computer Vision}, 202--221. Springer.

\bibitem[{Chen et~al.(2021{\natexlab{b}})Chen, Wu, Nie{\ss}ner, and Chang}]{chen2021d3net}
Chen, D.~Z.; Wu, Q.; Nie{\ss}ner, M.; and Chang, A.~X. 2021{\natexlab{b}}.
\newblock D3Net: A Speaker-Listener Architecture for Semi-supervised Dense Captioning and Visual Grounding in RGB-D Scans.
\newblock \emph{arXiv preprint arXiv:2112.01551}.

\bibitem[{Chen et~al.(2024{\natexlab{a}})Chen, Chen, Zhang, Li, Yu, Fei, Zhu, Fan, and Chen}]{LL3da2024}
Chen, S.; Chen, X.; Zhang, C.; Li, M.; Yu, G.; Fei, H.; Zhu, H.; Fan, J.; and Chen, T. 2024{\natexlab{a}}.
\newblock LL3DA: Visual Interactive Instruction Tuning for Omni-3D Understanding, Reasoning, and Planning.
\newblock In \emph{CVPR}, 26418--26428.

\bibitem[{Chen et~al.(2022)Chen, Guhur, Tapaswi, Schmid, and Laptev}]{chen2022language}
Chen, S.; Guhur, P.-L.; Tapaswi, M.; Schmid, C.; and Laptev, I. 2022.
\newblock Language conditioned spatial relation reasoning for 3d object grounding.
\newblock In \emph{NIPS}.

\bibitem[{Chen et~al.(2023{\natexlab{b}})Chen, Zhu, Chen, Lei, Yu, and Chen}]{vote2cap2023}
Chen, S.; Zhu, H.; Chen, X.; Lei, Y.; Yu, G.; and Chen, T. 2023{\natexlab{b}}.
\newblock End-to-End 3D Dense Captioning with Vote2Cap-DETR.
\newblock In \emph{2023 IEEE/CVF Conference on Computer Vision and Pattern Recognition (CVPR)}, 11124–11133.

\bibitem[{Chen et~al.(2024{\natexlab{b}})Chen, Zhu, Li, Chen, Guo, Lei, Yu, Li, and Chen}]{vote2cap++2024}
Chen, S.; Zhu, H.; Li, M.; Chen, X.; Guo, P.; Lei, Y.; Yu, G.; Li, T.; and Chen, T. 2024{\natexlab{b}}.
\newblock Vote2Cap-DETR++: Decoupling Localization and Describing for End-to-End 3D Dense Captioning.
\newblock \emph{IEEE Transactions on Pattern Analysis and Machine Intelligence}, 46(11): 7331–7347.

\bibitem[{Chen et~al.(2025)Chen, Wu, Lei, Pollefeys, and Chen}]{chen2025compile}
Chen, Z.; Wu, J.; Lei, Z.; Pollefeys, M.; and Chen, C.~W. 2025.
\newblock Compile Scene Graphs with Reinforcement Learning.
\newblock \emph{arXiv preprint arXiv:2504.13617}.

\bibitem[{DeepSeek-AI(2025)}]{deepseekr12025}
DeepSeek-AI. 2025.
\newblock DeepSeek-R1: Incentivizing Reasoning Capability in LLMs via Reinforcement Learning.
\newblock \emph{arXiv preprint arXiv:2501.12948}.

\bibitem[{Deng et~al.(2025)Deng, He, Jiang, Wang, Dayoub, and Reid}]{deng20253dllava}
Deng, J.; He, T.; Jiang, L.; Wang, T.; Dayoub, F.; and Reid, I. 2025.
\newblock 3D-LLaVA: Towards Generalist 3D LMMs with Omni Superpoint Transformer.
\newblock In \emph{Proceedings of the IEEE/CVF Conference on Computer Vision and Pattern Recognition}.

\bibitem[{Feng et~al.(2021)Feng, Li, Li, Zhang, Zhang, Zhu, Zhang, Wang, and Mian}]{feng2021free}
Feng, M.; Li, Z.; Li, Q.; Zhang, L.; Zhang, X.; Zhu, G.; Zhang, H.; Wang, Y.; and Mian, A. 2021.
\newblock Free-form Description Guided 3D Visual Graph Network for Object Grounding in Point Cloud.
\newblock In \emph{2021 IEEE/CVF International Conference on Computer Vision (ICCV)}, 3702--3711.

\bibitem[{Fu et~al.(2025)Fu, Liu, Chen, Nie, and Xiong}]{scenellm2025}
Fu, R.; Liu, J.; Chen, X.; Nie, Y.; and Xiong, W. 2025.
\newblock Scene-LLM: Extending Language Model for 3D Visual Reasoning.
\newblock In \emph{Proceedings of the Winter Conference on Applications of Computer Vision (WACV)}, 2195--2206.

\bibitem[{Halacheva et~al.(2025)Halacheva, Zaech, Wang, Paudel, and Gool}]{halacheva2025gaussianvlm}
Halacheva, A.-M.; Zaech, J.-N.; Wang, X.; Paudel, D.~P.; and Gool, L.~V. 2025.
\newblock GaussianVLM: Scene-centric 3D Vision-Language Models using Language-aligned Gaussian Splats for Embodied Reasoning and Beyond.
\newblock \emph{arXiv preprint arXiv:2507.00886}.

\bibitem[{He et~al.(2021)He, Zhao, Luo, Hui, Huang, Zhang, and Liu}]{transrefer3d}
He, D.; Zhao, Y.; Luo, J.; Hui, T.; Huang, S.; Zhang, A.; and Liu, S. 2021.
\newblock TransRefer3D: Entity-and-Relation Aware Transformer for Fine-Grained 3D Visual Grounding.
\newblock In \emph{Proceedings of the 29th ACM International Conference on Multimedia}.

\bibitem[{Hong et~al.(2023)Hong, Zhen, Chen, Zheng, Du, Chen, and Gan}]{hong2023dllm}
Hong, Y.; Zhen, H.; Chen, P.; Zheng, S.; Du, Y.; Chen, Z.; and Gan, C. 2023.
\newblock 3D-{LLM}: Injecting the 3D World into Large Language Models.
\newblock In \emph{Thirty-seventh Conference on Neural Information Processing Systems}.

\bibitem[{Hu et~al.(2022)Hu, Shen, Wallis, Allen-Zhu, Li, Wang, Wang, and Chen}]{hu2022lora}
Hu, E.~J.; Shen, Y.; Wallis, P.; Allen-Zhu, Z.; Li, Y.; Wang, S.; Wang, L.; and Chen, W. 2022.
\newblock Lo{RA}: Low-Rank Adaptation of Large Language Models.
\newblock In \emph{International Conference on Learning Representations}.

\bibitem[{Huang et~al.(2024{\natexlab{a}})Huang, Chen, Wang, Huang, Xu, Wang, Liu, Cheng, Zhao, Pang, and Zhao}]{huang2024chatscene}
Huang, H.; Chen, Y.; Wang, Z.; Huang, R.; Xu, R.; Wang, T.; Liu, L.; Cheng, X.; Zhao, Y.; Pang, J.; and Zhao, Z. 2024{\natexlab{a}}.
\newblock Chat-Scene: Bridging 3D Scene and Large Language Models with Object Identifiers.
\newblock In \emph{The Thirty-eighth Annual Conference on Neural Information Processing Systems}.

\bibitem[{Huang et~al.(2024{\natexlab{b}})Huang, Yong, Ma, Linghu, Li, Wang, Li, Zhu, Jia, and Huang}]{huang2024an}
Huang, J.; Yong, S.; Ma, X.; Linghu, X.; Li, P.; Wang, Y.; Li, Q.; Zhu, S.-C.; Jia, B.; and Huang, S. 2024{\natexlab{b}}.
\newblock An Embodied Generalist Agent in 3D World.
\newblock In \emph{ICLR 2024 Workshop: How Far Are We From AGI}.

\bibitem[{Huang et~al.(2021)Huang, Lee, Chen, and Liu}]{huang2021text}
Huang, P.-H.; Lee, H.-H.; Chen, H.-T.; and Liu, T.-L. 2021.
\newblock Text-Guided Graph Neural Networks for Referring 3D Instance Segmentation.
\newblock \emph{Proceedings of the AAAI Conference on Artificial Intelligence}, 35(2): 1610--1618.

\bibitem[{Huang et~al.(2022)Huang, Chen, Jia, and Wang}]{huang2022multi}
Huang, S.; Chen, Y.; Jia, J.; and Wang, L. 2022.
\newblock Multi-View Transformer for 3D Visual Grounding.
\newblock In \emph{Proceedings of the IEEE/CVF Conference on Computer Vision and Pattern Recognition}, 15524--15533.

\bibitem[{Huang et~al.(2025{\natexlab{a}})Huang, Zhang, Wang, and Tang}]{huang20253d}
Huang, T.; Zhang, Z.; Wang, Y.; and Tang, H. 2025{\natexlab{a}}.
\newblock 3D CoCa: Contrastive Learners are 3D Captioners.
\newblock \emph{arXiv preprint arXiv:2504.09518}.

\bibitem[{Huang et~al.(2025{\natexlab{b}})Huang, Zhang, Zhang, and Zhao}]{huang2025dc}
Huang, T.; Zhang, Z.; Zhang, R.; and Zhao, Y. 2025{\natexlab{b}}.
\newblock DC-Scene: Data-Centric Learning for 3D Scene Understanding.
\newblock \emph{arXiv preprint arXiv:2505.15232}.

\bibitem[{Jain et~al.(2022)Jain, Gkanatsios, Mediratta, and Fragkiadaki}]{jain2022bottom}
Jain, A.; Gkanatsios, N.; Mediratta, I.; and Fragkiadaki, K. 2022.
\newblock Bottom up top down detection transformers for language grounding in images and point clouds.
\newblock In \emph{ECCV}, 417--433. Springer.

\bibitem[{Jia et~al.(2024)Jia, Chen, Yu, Wang, Niu, Liu, Li, and Huang}]{jia2024sceneverse}
Jia, B.; Chen, Y.; Yu, H.; Wang, Y.; Niu, X.; Liu, T.; Li, Q.; and Huang, S. 2024.
\newblock Sceneverse: Scaling 3d vision-language learning for grounded scene understanding.
\newblock In \emph{European Conference on Computer Vision (ECCV)}.

\bibitem[{Jiao et~al.(2022)Jiao, Chen, Jie, Chen, Ma, and Jiang}]{MORE_2022}
Jiao, Y.; Chen, S.; Jie, Z.; Chen, J.; Ma, L.; and Jiang, Y.-G. 2022.
\newblock MORE: Multi-Order RElation Mining for Dense Captioning in 3D Scenes.
\newblock In \emph{In Proceedings of the European conference on computer vision}, 528–545.

\bibitem[{Jin et~al.(2023)Jin, Hayat, Yang, Guo, and Lei}]{jin2023-3D-VLP}
Jin, Z.; Hayat, M.; Yang, Y.; Guo, Y.; and Lei, Y. 2023.
\newblock Context-aware Alignment and Mutual Masking for 3D-Language Pre-training.
\newblock In \emph{Proceedings of the IEEE/CVF Conference on Computer Vision and Pattern Recognition}, 10984--10994.

\bibitem[{Kim et~al.(2025)Kim, Lim, Lee, Kim, and Kim}]{bica2025}
Kim, M.; Lim, H.; Lee, S.; Kim, B.; and Kim, G. 2025.
\newblock Bi-directional Contextual Attention for 3D Dense Captioning.
\newblock In \emph{In Proceedings of the European conference on computer vision}, 385–401.

\bibitem[{Lichen et~al.(2021)Lichen, Daigang, Lu, and Dong}]{zhao2021_3DVG_Transformer}
Lichen, Z.; Daigang, C.; Lu, S.; and Dong, X. 2021.
\newblock {3DVG-Transformer}: Relation modeling for visual grounding on point clouds.
\newblock In \emph{ICCV}, 2928--2937.

\bibitem[{Lin(2004)}]{rouge2004}
Lin, C.-Y. 2004.
\newblock {ROUGE}: A Package for Automatic Evaluation of Summaries.
\newblock In \emph{Text Summarization Branches Out}, 74--81. Barcelona, Spain: Association for Computational Linguistics.

\bibitem[{Luo et~al.(2022)Luo, Fu, Kong, Gao, Ren, Shen, Xia, and Liu}]{luo20223d}
Luo, J.; Fu, J.; Kong, X.; Gao, C.; Ren, H.; Shen, H.; Xia, H.; and Liu, S. 2022.
\newblock 3D-SPS: Single-Stage 3D Visual Grounding via Referred Point Progressive Selection.
\newblock \emph{arXiv preprint arXiv:2204.06272}.

\bibitem[{Luo, Johnson, and Lee(2024)}]{luo2024view}
Luo, T.; Johnson, J.; and Lee, H. 2024.
\newblock View selection for 3d captioning via diffusion ranking.
\newblock In \emph{European Conference on Computer Vision}, 180--197. Springer.

\bibitem[{Luo et~al.(2023)Luo, Rockwell, Lee, and Johnson}]{luo2023scalable}
Luo, T.; Rockwell, C.; Lee, H.; and Johnson, J. 2023.
\newblock Scalable 3D Captioning with Pretrained Models.
\newblock In \emph{Thirty-seventh Conference on Neural Information Processing Systems Datasets and Benchmarks Track}.

\bibitem[{Ma et~al.(2023)Ma, Yong, Zheng, Li, Liang, Zhu, and Huang}]{ma2022sqa3d}
Ma, X.; Yong, S.; Zheng, Z.; Li, Q.; Liang, Y.; Zhu, S.-C.; and Huang, S. 2023.
\newblock SQA3D: Situated Question Answering in 3D Scenes.
\newblock In \emph{International Conference on Learning Representations}.

\bibitem[{Mao et~al.(2023)Mao, Yang, Chen, Yi, and Liu}]{mao2023REMAN}
Mao, A.; Yang, Z.; Chen, W.; Yi, R.; and Liu, Y.-j. 2023.
\newblock Complete 3D Relationships Extraction Modality Alignment Network for 3D Dense Captioning.
\newblock \emph{IEEE Transactions on Visualization and Computer Graphics}.

\bibitem[{Mo and Liu(2024)}]{bridgeqa2024}
Mo, W.; and Liu, Y. 2024.
\newblock Bridging the gap between 2D and 3D visual question answering: a fusion approach for 3D VQA.
\newblock In \emph{Proceedings of the Thirty-Eighth AAAI Conference on Artificial Intelligence and Thirty-Sixth Conference on Innovative Applications of Artificial Intelligence and Fourteenth Symposium on Educational Advances in Artificial Intelligence}.

\bibitem[{OpenAI et~al.(2024)OpenAI, Achiam, Adler, Agarwal, Ahmad, Akkaya, Aleman, Almeida, Altenschmidt, Altman, Anadkat, Avila, Babuschkin, Balaji, Balcom, Baltescu, Bao, Bavarian, Belgum, Bello, Berdine, Bernadett-Shapiro, Berner, Bogdonoff, Boiko, Boyd, Brakman, Brockman, Brooks, Brundage, Button, Cai, Campbell, Cann, Carey, Carlson, Carmichael, Chan, Chang, Chantzis, Chen, Chen, Chen, Chen, Chen, Chess, Cho, Chu, Chung, Cummings, Currier, Dai, Decareaux, Degry, Deutsch, Deville, Dhar, Dohan, Dowling, Dunning, Ecoffet, Eleti, Eloundou, Farhi, Fedus, Felix, Fishman, Forte, Fulford, Gao, Georges, Gibson, Goel, Gogineni, Goh, Gontijo-Lopes, Gordon, Grafstein, Gray, Greene, Gross, Gu, Guo, Hallacy, Han, Harris, He, Heaton, Heidecke, Hesse, Hickey, Hickey, Hoeschele, Houghton, Hsu, Hu, Hu, Huizinga, Jain, Jain, Jang, Jiang, Jiang, Jin, Jin, Jomoto, Jonn, Jun, Kaftan, Łukasz Kaiser, Kamali, Kanitscheider, Keskar, Khan, Kilpatrick, Kim, Kim, Kim, Kirchner, Kiros, Knight, Kokotajlo, Łukasz Kondraciuk,
  Kondrich, Konstantinidis, Kosic, Krueger, Kuo, Lampe, Lan, Lee, Leike, Leung, Levy, Li, Lim, Lin, Lin, Litwin, Lopez, Lowe, Lue, Makanju, Malfacini, Manning, Markov, Markovski, Martin, Mayer, Mayne, McGrew, McKinney, McLeavey, McMillan, McNeil, Medina, Mehta, Menick, Metz, Mishchenko, Mishkin, Monaco, Morikawa, Mossing, Mu, Murati, Murk, Mély, Nair, Nakano, Nayak, Neelakantan, Ngo, Noh, Ouyang, O'Keefe, Pachocki, Paino, Palermo, Pantuliano, Parascandolo, Parish, Parparita, Passos, Pavlov, Peng, Perelman, de~Avila Belbute~Peres, Petrov, de~Oliveira~Pinto, Michael, Pokorny, Pokrass, Pong, Powell, Power, Power, Proehl, Puri, Radford, Rae, Ramesh, Raymond, Real, Rimbach, Ross, Rotsted, Roussez, Ryder, Saltarelli, Sanders, Santurkar, Sastry, Schmidt, Schnurr, Schulman, Selsam, Sheppard, Sherbakov, Shieh, Shoker, Shyam, Sidor, Sigler, Simens, Sitkin, Slama, Sohl, Sokolowsky, Song, Staudacher, Such, Summers, Sutskever, Tang, Tezak, Thompson, Tillet, Tootoonchian, Tseng, Tuggle, Turley, Tworek, Uribe, Vallone,
  Vijayvergiya, Voss, Wainwright, Wang, Wang, Wang, Ward, Wei, Weinmann, Welihinda, Welinder, Weng, Weng, Wiethoff, Willner, Winter, Wolrich, Wong, Workman, Wu, Wu, Wu, Xiao, Xu, Yoo, Yu, Yuan, Zaremba, Zellers, Zhang, Zhang, Zhao, Zheng, Zhuang, Zhuk, and Zoph}]{openai2024gpt4technicalreport}
OpenAI; Achiam, J.; Adler, S.; Agarwal, S.; Ahmad, L.; Akkaya, I.; Aleman, F.~L.; Almeida, D.; Altenschmidt, J.; Altman, S.; Anadkat, S.; Avila, R.; Babuschkin, I.; Balaji, S.; Balcom, V.; Baltescu, P.; Bao, H.; Bavarian, M.; Belgum, J.; Bello, I.; Berdine, J.; Bernadett-Shapiro, G.; Berner, C.; Bogdonoff, L.; Boiko, O.; Boyd, M.; Brakman, A.-L.; Brockman, G.; Brooks, T.; Brundage, M.; Button, K.; Cai, T.; Campbell, R.; Cann, A.; Carey, B.; Carlson, C.; Carmichael, R.; Chan, B.; Chang, C.; Chantzis, F.; Chen, D.; Chen, S.; Chen, R.; Chen, J.; Chen, M.; Chess, B.; Cho, C.; Chu, C.; Chung, H.~W.; Cummings, D.; Currier, J.; Dai, Y.; Decareaux, C.; Degry, T.; Deutsch, N.; Deville, D.; Dhar, A.; Dohan, D.; Dowling, S.; Dunning, S.; Ecoffet, A.; Eleti, A.; Eloundou, T.; Farhi, D.; Fedus, L.; Felix, N.; Fishman, S.~P.; Forte, J.; Fulford, I.; Gao, L.; Georges, E.; Gibson, C.; Goel, V.; Gogineni, T.; Goh, G.; Gontijo-Lopes, R.; Gordon, J.; Grafstein, M.; Gray, S.; Greene, R.; Gross, J.; Gu, S.~S.; Guo, Y.; Hallacy,
  C.; Han, J.; Harris, J.; He, Y.; Heaton, M.; Heidecke, J.; Hesse, C.; Hickey, A.; Hickey, W.; Hoeschele, P.; Houghton, B.; Hsu, K.; Hu, S.; Hu, X.; Huizinga, J.; Jain, S.; Jain, S.; Jang, J.; Jiang, A.; Jiang, R.; Jin, H.; Jin, D.; Jomoto, S.; Jonn, B.; Jun, H.; Kaftan, T.; Łukasz Kaiser; Kamali, A.; Kanitscheider, I.; Keskar, N.~S.; Khan, T.; Kilpatrick, L.; Kim, J.~W.; Kim, C.; Kim, Y.; Kirchner, J.~H.; Kiros, J.; Knight, M.; Kokotajlo, D.; Łukasz Kondraciuk; Kondrich, A.; Konstantinidis, A.; Kosic, K.; Krueger, G.; Kuo, V.; Lampe, M.; Lan, I.; Lee, T.; Leike, J.; Leung, J.; Levy, D.; Li, C.~M.; Lim, R.; Lin, M.; Lin, S.; Litwin, M.; Lopez, T.; Lowe, R.; Lue, P.; Makanju, A.; Malfacini, K.; Manning, S.; Markov, T.; Markovski, Y.; Martin, B.; Mayer, K.; Mayne, A.; McGrew, B.; McKinney, S.~M.; McLeavey, C.; McMillan, P.; McNeil, J.; Medina, D.; Mehta, A.; Menick, J.; Metz, L.; Mishchenko, A.; Mishkin, P.; Monaco, V.; Morikawa, E.; Mossing, D.; Mu, T.; Murati, M.; Murk, O.; Mély, D.; Nair, A.; Nakano, R.;
  Nayak, R.; Neelakantan, A.; Ngo, R.; Noh, H.; Ouyang, L.; O'Keefe, C.; Pachocki, J.; Paino, A.; Palermo, J.; Pantuliano, A.; Parascandolo, G.; Parish, J.; Parparita, E.; Passos, A.; Pavlov, M.; Peng, A.; Perelman, A.; de~Avila Belbute~Peres, F.; Petrov, M.; de~Oliveira~Pinto, H.~P.; Michael; Pokorny; Pokrass, M.; Pong, V.~H.; Powell, T.; Power, A.; Power, B.; Proehl, E.; Puri, R.; Radford, A.; Rae, J.; Ramesh, A.; Raymond, C.; Real, F.; Rimbach, K.; Ross, C.; Rotsted, B.; Roussez, H.; Ryder, N.; Saltarelli, M.; Sanders, T.; Santurkar, S.; Sastry, G.; Schmidt, H.; Schnurr, D.; Schulman, J.; Selsam, D.; Sheppard, K.; Sherbakov, T.; Shieh, J.; Shoker, S.; Shyam, P.; Sidor, S.; Sigler, E.; Simens, M.; Sitkin, J.; Slama, K.; Sohl, I.; Sokolowsky, B.; Song, Y.; Staudacher, N.; Such, F.~P.; Summers, N.; Sutskever, I.; Tang, J.; Tezak, N.; Thompson, M.~B.; Tillet, P.; Tootoonchian, A.; Tseng, E.; Tuggle, P.; Turley, N.; Tworek, J.; Uribe, J. F.~C.; Vallone, A.; Vijayvergiya, A.; Voss, C.; Wainwright, C.; Wang,
  J.~J.; Wang, A.; Wang, B.; Ward, J.; Wei, J.; Weinmann, C.; Welihinda, A.; Welinder, P.; Weng, J.; Weng, L.; Wiethoff, M.; Willner, D.; Winter, C.; Wolrich, S.; Wong, H.; Workman, L.; Wu, S.; Wu, J.; Wu, M.; Xiao, K.; Xu, T.; Yoo, S.; Yu, K.; Yuan, Q.; Zaremba, W.; Zellers, R.; Zhang, C.; Zhang, M.; Zhao, S.; Zheng, T.; Zhuang, J.; Zhuk, W.; and Zoph, B. 2024.
\newblock GPT-4 Technical Report.
\newblock \emph{arXiv preprint arXiv:2303.08774}.

\bibitem[{Papineni et~al.(2002)Papineni, Roukos, Ward, and Zhu}]{bleu2002}
Papineni, K.; Roukos, S.; Ward, T.; and Zhu, W.-J. 2002.
\newblock BLEU: a method for automatic evaluation of machine translation.
\newblock In \emph{Proceedings of the 40th Annual Meeting on Association for Computational Linguistics}, ACL '02, 311–318. USA: Association for Computational Linguistics.

\bibitem[{Parelli et~al.(2023)Parelli, Delitzas, Hars, Vlassis, Anagnostidis, Bachmann, and Hofmann}]{parelli2023clip-guided}
Parelli, M.; Delitzas, A.; Hars, N.; Vlassis, G.; Anagnostidis, S.; Bachmann, G.; and Hofmann, T. 2023.
\newblock CLIP-Guided Vision-Language Pre-training for Question Answering in 3D Scenes.
\newblock In \emph{Proceedings of the IEEE/CVF Conference on Computer Vision and Pattern Recognition}, 5606--5611.

\bibitem[{Park et~al.(2025)Park, Kim, Kim, Kim, and Ro}]{dipr12025}
Park, S.; Kim, H.; Kim, J.; Kim, S.; and Ro, Y.~M. 2025.
\newblock DIP-R1: Deep Inspection and Perception with RL Looking Through and Understanding Complex Scenes.
\newblock \emph{arXiv preprint arXiv:2505.23179}.

\bibitem[{Poole et~al.(2023)Poole, Jain, Barron, and Mildenhall}]{cliprpercision}
Poole, B.; Jain, A.; Barron, J.~T.; and Mildenhall, B. 2023.
\newblock DreamFusion: Text-to-3D using 2D Diffusion.
\newblock In \emph{ICLR}.

\bibitem[{Qi et~al.(2025)Qi, Zhang, Fang, Wang, and Zhao}]{GPT4Scene}
Qi, Z.; Zhang, Z.; Fang, Y.; Wang, J.; and Zhao, H. 2025.
\newblock GPT4Scene: Understand 3D Scenes from Videos with Vision-Language Models.
\newblock \emph{arXiv:2501.01428}.

\bibitem[{Shao et~al.(2024)Shao, Wang, ihao Zhu, Xu, Song, Zhang, Y.K.~Li, and Guo}]{deepseek-math}
Shao, Z.; Wang, P.; ihao Zhu; Xu, R.; Song, J.; Zhang, M.; Y.K.~Li, Y.~W.; and Guo, D. 2024.
\newblock DeepSeekMath: Pushing the Limits of Mathematical Reasoning in Open Language Models.
\newblock \emph{CoRR}, abs/2402.03300.

\bibitem[{Song et~al.(2025)Song, Blukis, Tremblay, Tyree, Su, and Birchfield}]{song2025robospatial}
Song, C.~H.; Blukis, V.; Tremblay, J.; Tyree, S.; Su, Y.; and Birchfield, S. 2025.
\newblock RoboSpatial: Teaching Spatial Understanding to 2D and 3D Vision-Language Models for Robotics.
\newblock In \emph{7th Robot Learning Workshop: Towards Robots with Human-Level Abilities}.

\bibitem[{Tang et~al.(2024)Tang, Han, Li, Yu, Hao, Hu, and Chen}]{minigpt3d2024}
Tang, Y.; Han, X.; Li, X.; Yu, Q.; Hao, Y.; Hu, L.; and Chen, M. 2024.
\newblock MiniGPT-3D: Efficiently Aligning 3D Point Clouds with Large Language Models using 2D Priors.
\newblock In \emph{Proceedings of the 32nd ACM International Conference on Multimedia}, 6617–6626.

\bibitem[{Team et~al.(2025)Team, Modi, Veerubhotla, Rysbek, Huber, Anand, Bhoopchand, Wiltshire, Gillick, Kasenberg, Sgouritsa, Elidan, Liu, Winnemoeller, Jurenka, Cohan, She, Wilkowski, Alarakyia, McKee, Singh, Wang, Kunesch, Pîslar, Efron, Mahmoudieh, Kamienny, Wiltberger, Mohamed, Agarwal, Phal, Lee, Strinopoulos, Ko, Gold-Zamir, Haramaty, and Assael}]{geminipro2025}
Team, L.; Modi, A.; Veerubhotla, A.~S.; Rysbek, A.; Huber, A.; Anand, A.; Bhoopchand, A.; Wiltshire, B.; Gillick, D.; Kasenberg, D.; Sgouritsa, E.; Elidan, G.; Liu, H.; Winnemoeller, H.; Jurenka, I.; Cohan, J.; She, J.; Wilkowski, J.; Alarakyia, K.; McKee, K.~R.; Singh, K.; Wang, L.; Kunesch, M.; Pîslar, M.; Efron, N.; Mahmoudieh, P.; Kamienny, P.-A.; Wiltberger, S.; Mohamed, S.; Agarwal, S.; Phal, S.~M.; Lee, S.~J.; Strinopoulos, T.; Ko, W.-J.; Gold-Zamir, Y.; Haramaty, Y.; and Assael, Y. 2025.
\newblock Evaluating Gemini in an arena for learning.
\newblock \emph{arXiv preprint arXiv:2505.24477}.

\bibitem[{Tschannen et~al.(2025)Tschannen, Gritsenko, Wang, Naeem, Alabdulmohsin, Parthasarathy, Evans, Beyer, Xia, Mustafa, Hénaff, Harmsen, Steiner, and Zhai}]{siglip2}
Tschannen, M.; Gritsenko, A.; Wang, X.; Naeem, M.~F.; Alabdulmohsin, I.; Parthasarathy, N.; Evans, T.; Beyer, L.; Xia, Y.; Mustafa, B.; Hénaff, O.; Harmsen, J.; Steiner, A.; and Zhai, X. 2025.
\newblock SigLIP 2: Multilingual Vision-Language Encoders with Improved Semantic Understanding, Localization, and Dense Features.
\newblock \emph{arXiv preprint arXiv:2502.14786}.

\bibitem[{Vedantam, Zitnick, and Parikh(2015)}]{cider2015}
Vedantam, R.; Zitnick, C.~L.; and Parikh, D. 2015.
\newblock CIDEr: Consensus-based image description evaluation.
\newblock In \emph{2015 IEEE Conference on Computer Vision and Pattern Recognition (CVPR)}, 4566--4575.

\bibitem[{Wang et~al.(2022)Wang, Zhang, Yu, and Cai}]{spa2cap2022}
Wang, H.; Zhang, C.; Yu, J.; and Cai, W. 2022.
\newblock Spatiality-guided Transformer for 3D Dense Captioning on Point Clouds.
\newblock In \emph{Proceedings of the Thirty-First International Joint Conference on Artificial Intelligence}, 1393–1400.

\bibitem[{Wang et~al.(2025)Wang, Li, Xu, Qi, Yang, Ma, Liu, and Zhang}]{spatial3dllm}
Wang, X.; Li, Z.; Xu, Y.; Qi, J.; Yang, Z.; Ma, R.; Liu, X.; and Zhang, C. 2025.
\newblock Spatial 3D-LLM: Exploring Spatial Awareness in 3D Vision-Language Models.
\newblock \emph{arXiv preprint arXiv:2507.16524}.

\bibitem[{Wei et~al.(2022)Wei, Wang, Schuurmans, Bosma, brian ichter, Xia, Chi, Le, and Zhou}]{wei2022chain}
Wei, J.; Wang, X.; Schuurmans, D.; Bosma, M.; brian ichter; Xia, F.; Chi, E.~H.; Le, Q.~V.; and Zhou, D. 2022.
\newblock Chain of Thought Prompting Elicits Reasoning in Large Language Models.
\newblock In Oh, A.~H.; Agarwal, A.; Belgrave, D.; and Cho, K., eds., \emph{Advances in Neural Information Processing Systems}.

\bibitem[{Wu et~al.(2024)Wu, Jiang, Wang, Liu, Liu, Qiao, Ouyang, He, and Zhao}]{wu2024ptv3}
Wu, X.; Jiang, L.; Wang, P.-S.; Liu, Z.; Liu, X.; Qiao, Y.; Ouyang, W.; He, T.; and Zhao, H. 2024.
\newblock Point Transformer V3: Simpler, Faster, Stronger.
\newblock In \emph{CVPR}.

\bibitem[{Wu et~al.(2023)Wu, Cheng, Zhang, Cheng, and Zhang}]{wu2023eda}
Wu, Y.; Cheng, X.; Zhang, R.; Cheng, Z.; and Zhang, J. 2023.
\newblock Eda: Explicit text-decoupling and dense alignment for 3d visual grounding.
\newblock In \emph{CVPR}, 19231--19242.

\bibitem[{Xu et~al.(2024)Xu, Wang, Wang, Chen, Pang, and Lin}]{xu2024pointllm}
Xu, R.; Wang, X.; Wang, T.; Chen, Y.; Pang, J.; and Lin, D. 2024.
\newblock PointLLM: Empowering Large Language Models to Understand Point Clouds.
\newblock In \emph{ECCV}.

\bibitem[{Yang et~al.(2024)Yang, Kang, Huang, Zhao, Xu, Feng, and Zhao}]{yang2024depthv2}
Yang, L.; Kang, B.; Huang, Z.; Zhao, Z.; Xu, X.; Feng, J.; and Zhao, H. 2024.
\newblock Depth Anything V2.
\newblock \emph{arXiv preprint arXiv:2406.09414}.

\bibitem[{Yang et~al.(2021)Yang, Zhang, Wang, and Luo}]{yang2021sat}
Yang, Z.; Zhang, S.; Wang, L.; and Luo, J. 2021.
\newblock SAT: 2D Semantics Assisted Training for 3D Visual Grounding.
\newblock In \emph{ICCV}.

\bibitem[{Yuan et~al.(2025)Yuan, Jiang, Feng, Zhang, Cui, Li, and Zhao}]{scener12025}
Yuan, Z.; Jiang, S.; Feng, C.-M.; Zhang, Y.; Cui, S.; Li, Z.; and Zhao, N. 2025.
\newblock Scene-R1: Video-Grounded Large Language Models for 3D Scene Reasoning without 3D Annotations.
\newblock \emph{arXiv preprint arXiv:2506.17545}.

\bibitem[{Yuan et~al.(2021)Yuan, Yan, Liao, Zhang, Li, and Cui}]{yuan2021instancerefer}
Yuan, Z.; Yan, X.; Liao, Y.; Zhang, R.; Li, Z.; and Cui, S. 2021.
\newblock Instancerefer: Cooperative holistic understanding for visual grounding on point clouds through instance multi-level contextual referring.
\newblock In \emph{Proceedings of the IEEE/CVF International Conference on Computer Vision}, 1791--1800.

\bibitem[{Zhao et~al.(2024)Zhao, Lin, Ye, Pang, and Lau}]{zhao2024openscan}
Zhao, Y.; Lin, J.; Ye, S.; Pang, Q.; and Lau, R.~W. 2024.
\newblock OpenScan: A Benchmark for Generalized Open-Vocabulary 3D Scene Understanding.
\newblock \emph{arXiv preprint arXiv:2408.11030}.

\bibitem[{Zheng, Huang, and Wang(2025)}]{video3dllm}
Zheng, D.; Huang, S.; and Wang, L. 2025.
\newblock {Video-3D LLM}: Learning Position-Aware Video Representation for {3D} Scene Understanding.
\newblock In \emph{Proceedings of the IEEE/CVF Conference on Computer Vision and Pattern Recognition (CVPR)}.

\bibitem[{Zhi et~al.(2024)Zhi, Chen, Li, Ma, Sun, Xiang, Lei, Tan, and Gan}]{zhi2024lscenellm}
Zhi, H.; Chen, P.; Li, J.; Ma, S.; Sun, X.; Xiang, T.; Lei, Y.; Tan, M.; and Gan, C. 2024.
\newblock LSceneLLM: Enhancing Large 3D Scene Understanding Using Adaptive Visual Preferences.
\newblock \emph{arXiv preprint arXiv:2412.01292}.

\bibitem[{Zhong et~al.(2022)Zhong, Xu, Luo, and Ma}]{zhong2022contextual3DdenseCap}
Zhong, Y.; Xu, L.; Luo, J.; and Ma, L. 2022.
\newblock Contextual Modeling for 3D Dense Captioning on Point Clouds.
\newblock \emph{arXiv preprint arXiv:2210.03925}.

\bibitem[{Zhu et~al.(2024)Zhu, Wang, Zhang, Pang, and Liu}]{zhu2024llava}
Zhu, C.; Wang, T.; Zhang, W.; Pang, J.; and Liu, X. 2024.
\newblock LLaVA-3D: A Simple yet Effective Pathway to Empowering LMMs with 3D-awareness.
\newblock \emph{CoRR}, abs/2409.18125.

\bibitem[{Zhu et~al.(2023)Zhu, Ma, Chen, Deng, Huang, and Li}]{zhu20233d-vista}
Zhu, Z.; Ma, X.; Chen, Y.; Deng, Z.; Huang, S.; and Li, Q. 2023.
\newblock 3D-VisTA: Pre-trained Transformer for 3D Vision and Text Alignment.
\newblock In \emph{Proceedings of the IEEE/CVF International Conference on Computer Vision}, 2911--2921.

\end{thebibliography}
\end{document}